\documentclass{article}

\PassOptionsToPackage{numbers, compress}{natbib}

\usepackage[final]{neurips_2024}

\usepackage[utf8]{inputenc} %
\usepackage[T1]{fontenc}    %
\usepackage{hyperref}       %
\hypersetup{
    colorlinks,
    citecolor=blue,
    filecolor=black,
    linkcolor=blue,
    urlcolor=blue
}
\usepackage{url}            %
\usepackage{booktabs}       %
\usepackage{amsfonts}       %
\usepackage{nicefrac}       %
\usepackage{microtype}      %
\usepackage{xcolor}         %
\usepackage{xspace}
\usepackage{listings}
\usepackage{subcaption}
\usepackage{inconsolata}
\usepackage[inline]{enumitem}
\usepackage[font=small,labelfont=bf]{caption}
\usepackage{tikz}
\usepackage{wrapfig}
\usetikzlibrary{arrows.meta, positioning}

\definecolor{codegreen}{rgb}{0,0.6,0}
\definecolor{codegray}{rgb}{0.5,0.5,0.5}
\definecolor{codepurple}{rgb}{0.58,0,0.82}

\definecolor{backcolour}{RGB}{245,248,250}
\definecolor{emph}{RGB}{166,88,53}
\definecolor{nightblue}{RGB}{9,49,105}
\definecolor{keywords}{RGB}{207,33,46}
\definecolor{lightpurple}{RGB}{130,81,223}

\definecolor{skyblue}{RGB}{86,168,245}
\definecolor{deepblue}{rgb}{0,0,0.5}
\definecolor{deepred}{rgb}{0.6,0,0}
\definecolor{deepgreen}{rgb}{0,0.5,0}

\lstdefinestyle{mystyle}{
    backgroundcolor=\color{backcolour},    %
    commentstyle=\color{codegreen},
    keywordstyle=\color{keywords},
    stringstyle=\color{nightblue},
    basicstyle=\fontsize{7}{8}\ttfamily,
    breakatwhitespace=true,         
    breaklines=true,                 
    captionpos=b,                    
    keepspaces=true,                 
    numberstyle=\tiny\color{codegray},
    numbersep=2pt,                  
    showspaces=false,                
    showstringspaces=false,
    showtabs=false,                  
    tabsize=2,
    emph={bundle,@trace,trace,backward, instruction, code, documentation, variables, constraints,inputs,others,outputs,feedback,actual_problem_instance,variable1_name,variable2_name,variable1_value1,variable1_value2,variable2_value1,variable2_value2,feedback_1,feedback_2,node,@trace_class},
    emphstyle={\color{codepurple}},
    linewidth=1\columnwidth,
    frame=tb,    
    xrightmargin=0pt,
    xleftmargin=0.23cm,
    numbers=left,
    aboveskip=0.2cm,
    belowskip=0.1cm,
}

\usepackage{amsmath}
\usepackage{amsfonts}
\usepackage{amssymb}
\usepackage{amsthm}
\usepackage{bm}
\usepackage{bbm}
\usepackage{mathtools}
\usepackage{enumitem}
\usepackage{thmtools,thm-restate}
\usepackage{algorithm}
\usepackage[noend]{algorithmic}
\usepackage[capitalise]{cleveref}
\usepackage{color}
\usepackage{comment}

\theoremstyle{plain}
\newtheorem{theorem}{Theorem}%

\theoremstyle{definition}
\newtheorem{example}{Example}%
\newtheorem{assumption}{Assumption}%
\newtheorem{axiom}{Axiom}%

\def\TT{\mathcal{T}}
\def\XX{\mathcal{X}}

\newcommand{\gray}[1]{\textcolor{gray}{#1}}

\title{%

Trace is the Next AutoDiff: Generative Optimization with Rich Feedback, Execution Traces, and LLMs
}

\author{%
  Ching-An Cheng\thanks{Equal contribution} \\
  Microsoft Research\\
  \texttt{chinganc@microsoft.com} \\
  \And
  Allen Nie$^*$ \\
  Stanford \\
  \texttt{anie@cs.stanford.edu} \\
  \And
  Adith Swaminathan$^*$ \\
  Netflix \\
  \texttt{aswaminathan@netflix.com} \\
}

\newcommand{\opto}{OPTO\xspace}
\newcommand{\optolong}{Optimization with Trace Oracle\xspace}
\newcommand{\algo}{OptoPrime\xspace}
\newcommand{\trace}{Trace\xspace}

\def\node{\texttt{node}\xspace}
\def\bundle{\texttt{bundle}\xspace}
\def\trainable{\texttt{trainable}\xspace}
\def\backward{\texttt{backward}\xspace}
\def\propagate{\texttt{propagate}\xspace}
\def\step{\texttt{step}\xspace}
\def\graphpropagator{MSP\xspace}
\def\exception{\texttt{TraceExecutionError}\xspace}
\newcommand{\rev}{}
\begin{document}

\maketitle

\begin{abstract}
We study a class of optimization problems motivated by automating the design and update of AI systems like coding assistants, robots, and copilots.
\rev{AutoDiff frameworks, like PyTorch, enable efficient end-to-end optimization of differentiable systems. However, }
general computational workflows can be non-differentiable and involve rich feedback (e.g. console output or user's responses), heterogeneous parameters (e.g. prompts, codes), and intricate objectives (beyond maximizing a score).
\rev{We investigate \emph{end-to-end generative optimization} -- using generative models such as LLMs within the optimizer for automatic updating of general computational workflows. %
We discover that workflow execution traces are akin to back-propagated gradients in AutoDiff and can provide key information to interpret feedback for efficient optimization.}
Formally, we frame a new mathematical setup, %
\optolong (\opto). %
In \opto, an optimizer receives an execution trace along with feedback on the computed output and updates parameters iteratively.
\rev{We provide a Python library, \trace,}
that efficiently converts a workflow optimization problem into an \opto instance using PyTorch-like syntax.
Using \trace, we develop a general LLM-based generative optimizer called \algo. %
In empirical studies, we find that \algo is capable of first-order numerical optimization, prompt optimization, hyper-parameter tuning, robot controller design, code debugging, etc., and is often competitive with specialized optimizers for each domain.
\rev{We envision \trace as an open research platform for devising novel generative optimizers and} %
developing the next generation of interactive learning agents. %
Website: \url{https://microsoft.github.io/Trace/}.

\end{abstract}

\section{Introduction}
\label{sec:intro}

Computational workflows that integrate large language models (LLMs), machine learning (ML) models, orchestration, retrievers, tools, etc., power many state-of-the-art AI applications~\citep{compound-ai-blog}: from chatbots~\citep{openai2023chatgpt}, coding assistants~\citep{GithubCopilot}, robots~\citep{liang2023code}, to multi-agent systems~\citep{wu2023autogen}.
However designing a computational workflow requires laborious engineering because many heterogeneous parameters (e.g. prompts, orchestration code, and ML hyper-parameters) are involved.
Moreover, after deployment any erroneous behaviors of the workflow persist unless a developer manually updates it.

We study a class of optimization problems motivated by automating the design and update of computational workflows. %
Computational workflows %
produce optimization problems with heterogeneous parameters, rich feedback (e.g. console output and user's verbal responses), and intricate objectives (beyond maximizing a score).
Moreover, a workflow %
can have interdependent steps (e.g. adaptive orchestration, feedback control loops) and/or involve non-differentiable, semi-black-box, stochastic operations (e.g. ML models, simulations) whose behavior cannot be succinctly captured. %
As a result, the structure of the computation may change as the parameters and the inputs of the workflow vary.

Due to its complexity, computational workflow optimization is usually framed as a black-box~\citep{conn2009introduction} or algorithm configuration~\citep{hutter2011sequential} problem, and is tackled by general techniques like Bayesian Optimization~\citep{frazier2018bayesian}, Evolutionary Algorithms~\citep{back1993overview}, Reinforcement Learning (RL)~\citep{ginebra1995response} using scalar scores as feedback.
But one observation of scalar feedback alone does not provide an improvement signal, so these algorithms are very inefficient when the parameter space is large (e.g. codes or natural language prompts).
Recently LLM-based optimizers~\citep{zhou2022large,pryzant2023automatic,yang2024large,schnabel2024prompts,zelikman2023self,nie2023importance} have been proposed as \emph{generative optimizers} to improve efficiency, leveraging the prior of generative models learned from large pre-training corpora to optimize complex prompts and codes.
\rev{In this paper, we investigate how \emph{generative optimization} can be applied more broadly and systematically to optimize a general computational workflow end-to-end.
\cref{sec:related} discusses related works in generative optimization. }

\vspace{-1mm}
\subsection{Toward Efficient End-to-End Optimization of Computational Workflows}
\vspace{-1mm}

Inspired by back-propagation~\citep{rumelhart1986learning}, we take an end-to-end approach to computational workflow optimization. %
AutoDiff frameworks~\citep{AutoDiff, paszke2019pytorch} have scaled the back-propagation algorithm to optimize differentiable workflows (e.g., neural networks) with billions of parameters.
We extend the idea of  AutoDiff and design a new framework, \trace, for jointly optimizing all of the heterogeneous parameters in general computational workflows, which may not be differentiable.

\begin{figure}[b]
    \centering
    \includegraphics[width=0.92\textwidth]{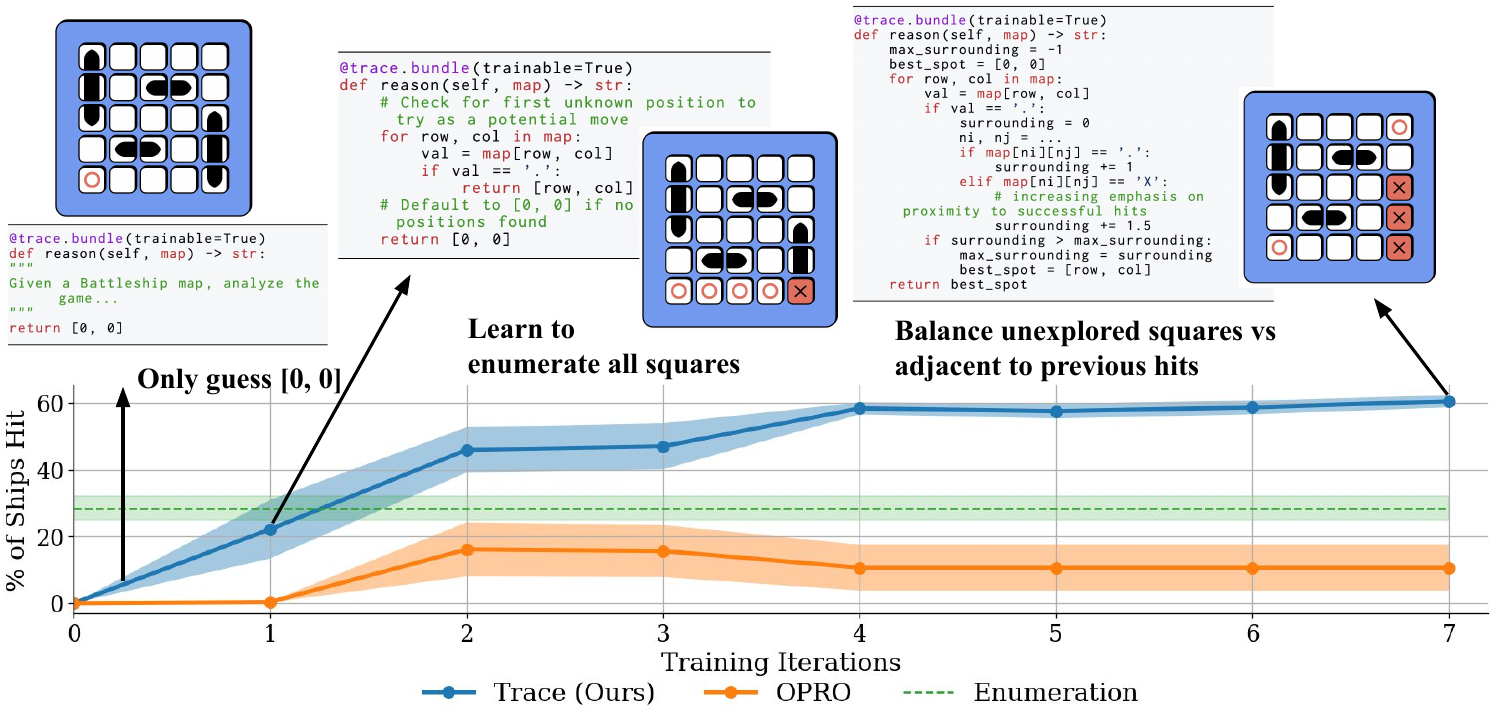}
    \caption{\textbf{Learning Example in Battleship.} %
    \trace automatically optimizes heterogeneous parameters (multiple codes) to train an agent to Battleship. %
    Means and standard errors are computed over 10 random seeds. %
    }
    \label{fig:battleship-learning}
\end{figure}

\begin{figure}
\centering
\begin{subfigure}[b]{0.35\textwidth}
\begin{lstlisting}[language=Python,breaklines=true,showstringspaces=false,basicstyle=\fontsize{6}{6}\ttfamily,numbers=none,
    escapechar=!
    ]
class Policy(trace.Module):

 def forward(self, map):
   plan = self.reason(map)
   output = self.act(map, plan)
   return output

 @trace.bundle(trainable=True)
 def reason(self, map) -> str:
   """
   Given a Battleship map, analyze the game...
   """
   return [0, 0]

 @trace.bundle(trainable=True)
 def act(self, map, plan):
   """
   Given a map and plan, select a target coordinate...
   """
   return
\end{lstlisting}
\caption{We write a trainable policy in Python using \trace operators. }
\label{fig:policy}
\end{subfigure}
~
\begin{subfigure}[b]{0.37\textwidth}
\begin{lstlisting}[language=Python,breaklines=true,showstringspaces=false,basicstyle=\fontsize{6}{6}\ttfamily,numbers=none,
    escapechar=!
    ]
policy = Policy()
params = policy.parameters()
optimizer = trace.Optimizer(params)

env = gym.make('Battleship-v0')
board = env.reset()
done, !feedback! = False, None
while not done:
  # Forward pass
  try:
    target = policy(board)
    board, !feedback!, done = env.step(target.data)
  except TraceExecutionError as e:
    !feedback! = str(e)
    target = e.exception_node
  # Backward pass and update
  optimizer.zero_feedback()
  optimizer.backward(target, !feedback!)
  optimizer.step()
\end{lstlisting}
\caption{We then use PyTorch-like optimization syntax to train the policy.}
\label{fig:opt loop}
\end{subfigure}
\begin{subfigure}[b]{0.22\textwidth}
\includegraphics[width=\textwidth]{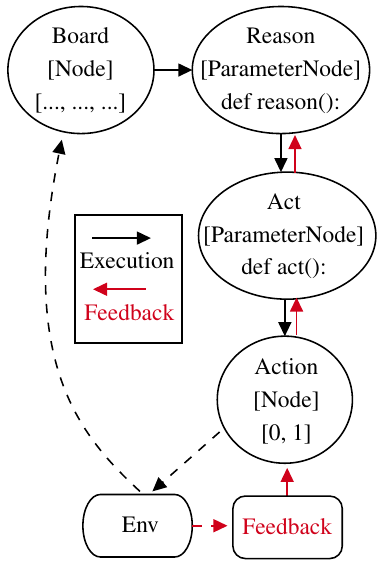}
\caption{\trace automatically records execution DAG.}
\label{fig:battleship graph}
\end{subfigure}
\caption{\textbf{Python Code of the Battleship Example.} To build a self-adapting agent with \trace, we only need to annotate some empty functions (\texttt{reason}, \texttt{act}) and set up an optimizer following PyTorch semantics. For space, we trim the docstrings of the empty functions with ``$\dots$'' and list them in~\cref{sec:optimized params examples}. \trace then builds a DAG as the workflow executes and updates the parameters (see~\cref{fig:battleship-learning} for the result).
}
\label{fig:overall_pipeline}
\vspace{-4mm}
\end{figure}

\trace treats a computational workflow as a graph like a neural network, where nodes are either inputs, parameters or the results of computation steps, and directional edges denote how nodes are created from others.
But, instead of gradients, \trace  propagates the \emph{execution trace} of a workflow (which records the intermediate computed results and how they are used), %
via the notion of minimal subgraph (see \cref{sec:backward}).
We show that propagating the execution trace subsumes back-propagation for differentiable workflows, and remains applicable even for non-differentiable workflows. %
Viewing a workflow as a graph and using its execution trace is standard for software engineering; for instance, human developers use such traces to debug distributed systems~\citep{nagaraj2012structured}. %
Our novel insight is that execution traces %
also enable end-to-end generative optimization, because they provide the information needed to relate rich feedback to the parameters in general workflows. %

\vspace{-1mm}
\subsection{Example of \trace in Action}
\vspace{-1mm}

\trace is available as a Python library with an API inspired by PyTorch~\citep{paszke2019pytorch}.
A user declares the parameters to be optimized using a \trainable flag,  decorates the workflow with \node and \bundle wrappers, and runs a \trace optimizer -- just like how they would declare and train neural networks.

Consider building an AI agent for the Battleship game (\cref{fig:battleship-learning}).
The agent's policy (\cref{fig:policy})  has two components (\verb|reason| and \verb|act|), which are chained together to react to different board configurations. The Battleship environment provides feedback (binary reward in texts) to tell if the agent's action hit the hidden ships, and the goal is to hit all hidden ships as fast as possible.
Consider how a human programmer might approach the problem. They may run the policy and change the code based on the observed feedback.
They may rewrite the code a few times to try different heuristics to solve this problem.
They will fix any execution errors (e.g., out-of-bounds exceptions) by using stack traces.

Our \trace framework accomplishes the programmer's goal automatically without adding complexity to the Python code. The user declares \verb|reason| and \verb|act| as trainable (\cref{fig:policy}) and then runs the agent in a PyTorch-like training loop (\cref{fig:opt loop}).
During the execution, \trace records a directed acyclic graph (DAG) (\cref{fig:battleship graph}) and uses it to compute the execution trace for optimization.
\trace also automatically catches errors (e.g., syntax/semantic errors) and can use them as feedback.
In~\cref{fig:battleship-learning}, we show what the agent learns as \trace optimizes\footnote{We use a new general-purpose LLM-based optimizer \algo that we detail in~\cref{sec:optimizer}.} its policy, where the learned policy is evaluated on new randomly generated games for generalization. %
The agent can quickly improve its performance and learn strategies that are increasingly complex.
\rev{Remarkably, there is no mention of the specific Battleship environment API, nor details on how the functions} \verb|reason| and \verb|act| \rev{should behave or adapt in \cref{fig:policy}.}
The \trace optimizer figures out all the details through interactions as the computational graph unfolds and the feedback on the output is observed.
Beyond code as parameters in this example, we also have experiments in~\cref{sec:expts} where prompts and other heterogenous parameters are optimized.

\vspace{-1mm}
\subsection{A New World of Optimization} %
\vspace{-1mm}

The design of \trace is based on a new mathematical setup of iterative optimization, which we call \optolong (\opto). In \opto, an optimizer selects parameters and receives a computational graph as well as feedback on the computed output.
\trace is a tool to %
efficiently convert the optimization of computational workflows into \opto problems in practice.

We argue that framing computational workflow optimization as \opto can lead to faster convergence than a black-box approach.
We present a constructive proof:  We design a general-purpose efficient generative optimizer called \algo, for \opto.
\algo turns \opto to a sequence of pseudo-algorithm problems.
In each iteration of \opto, we format the execution trace and output feedback %
as a pseudo-algorithm question and present it to an LLM for solution (GPT-4 using a ReAct-CoT prompt listed in~\cref{sec:algo_detail}).
In experiments, we apply \algo unchanged to many disparate applications like prompt optimization, first-order numerical optimization, hyper-parameter tuning, and robot controller design.
We find that the general purpose \algo is competitive with specialized optimizers for each domain, e.g., achieving $10\%$ higher accuracy on BigBenchHard~\citep{suzgun2022challenging} when optimizing a DSPy~\citep{khattab2023dspy} program compared to their hand-designed optimizer. %

\trace, \opto, and \algo together provide the first tractable algorithm for optimizing general computational workflows end-to-end.
The \trace framework
\begin{enumerate*}[label=\textit{\alph*)}]
    \item  leverages a workflow's structure and
    \item  can incorporate rich feedback beyond scores,
\end{enumerate*}
extending AutoDiff to complicated, non-differentiable computational workflows.
With \trace, we conjecture that ``training deep agent networks'' (which fluidly mix computation of tensors, LLMs, and other programmable tools) will soon be possible.

\section{Optimization with Trace Oracle}
\label{sec:opto}

\opto is the foundation of \trace. In this section, we define this graph-based abstraction of iterative optimization and discuss how \opto covers various computational workflow optimization problems.

\vspace{-2mm}
\paragraph{Preliminary}
We review the definition of a computational graph (see \cref{fig:battleship graph}). %
A computational graph $g$ is a DAG, where a node represents an object (such as tensors, strings, etc.) and an edge %
denotes an input-output relationship. %
We call a node without parents a root and a node without children a leaf, which are the inputs and outputs of the computational graph.
In the context of optimization, some inputs are marked as trainable \emph{parameters}, which are denoted as $\{X_\theta\}$.
For a node $X$, its parents are the inputs to an operator that creates $X$. The descendants of node $X$ are those that can be reached from $X$ following the directed edges; the ancestors are defined conversely.
Without loss of generality, we suppose that all computational operators have a unitary output\footnote{A multi-output operator can always be modeled by a single-output operator and single-output indexers.}.
In this way, we can associate the operator that creates the child node with the child node, %
and the full computation can be represented compactly as a DAG without explicitly representing the operators.
\rev{The execution trace of a computational workflow is defined as the sequence of operations and their execution results invoked when computing the output from a set of inputs; execution traces can be expressed as a computational graph as defined above.}

\subsection{Problem Definition of \opto}

\begin{wrapfigure}{o}{0.4\textwidth}
    \vspace{-5mm}
    \centering
    \includegraphics[width=0.4\textwidth]{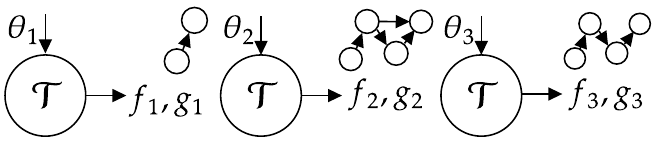}
    \caption{\textbf{Iterations of \opto.} When  $\theta\in\Theta$ is selected, the Trace Oracle $\TT$ returns trace feedback $\tau = (f,g)$, where $g$ is a computational graph using $\theta$ as input and $f$ is the feedback given to the output of $g$. }
    \label{fig:opto-diagram}
    \vspace{-3mm}
\end{wrapfigure}

\opto is an \emph{abstract} setup of iterative computational workflow optimization.  %
An \opto problem instance is defined by a tuple $(\Theta, \omega, \TT)$, where $\Theta$ is the parameter space, $\omega$ is the context of the problem, and $\TT$ is a Trace Oracle. %
In each iteration, the optimizer selects a parameter $\theta\in\Theta$, which can be heterogeneous. Then the Trace Oracle $\TT$ returns a \emph{trace feedback}, denoted as $\tau = (f,g)$, where $g$ is the execution trace represented as a DAG %
(the parameter is contained in the root nodes of $g$), and $f$ is the feedback provided to exactly one of the output nodes of $g$.  Finally, the optimizer uses the trace feedback $\tau$ to update the parameter according to the context $\omega$ and proceeds to the next iteration, as shown in~\cref{fig:opto-diagram}.

In \opto, the output feedback $f$ is generic, such as scores, gradients, hints/explanations expressed in natural language, and console messages.
The context $\omega$ provides invariant information to interpret the output feedback $f$ as well as any known side-information, e.g., desired properties of the parameters.
The context $\omega$ is fixed for an \opto problem instance (similar to an instruction, or a problem definition), whereas the output feedback $f$ can change with the parameter $\theta\in\Theta$ and the resulting computation $g$.
For example,  $\omega$ may be ``Minimize a loss function'' and $f$ is a loss. Alternatively, $\omega$ can be open-ended, like ``Follow the feedback'' and $f$ describes how an output should be changed.
In \cref{sec:constructing opto problem}, we discuss how to define the context and output feedback when constructing \opto problems in practice. %
In this paper, we focus on \opto problems where $f$ and $\omega$ can be expressed compactly in text. This covers a wide range of problems~\citep{cheng2023llf}, including those with scalar feedback. %

\opto differs from a black-box setup in that the execution trace $g$ shows the computational path toward the output, which provides information to construct a parameter update direction from $f$ and $\omega$.
In the minimization example above, when the execution trace $g$ is missing, it is unclear how the parameter can be improved given only a point evaluation of $f$. On the other hand, with $g$ documenting the functions used to create the output, an update direction (e.g., a gradient) can be derived. %
We highlight that the structure of the computational graph $g$ returned by the Trace Oracle $\TT$ can be different in each iteration in the general case (as in \cref{fig:opto-diagram}) because some workflows can change with different inputs and parameters.

To ground the \opto setup, we show how \opto is related to some existing problems with examples.
We discuss other examples like hyperparameter tuning and multi-agent systems in \cref{sec:more opto examples}.
\begin{example}[Neural network with back-propagation]
The parameters are the weights. $g$ is the neural computational graph and $f$ is the loss. An example context $\omega$ can be ``Minimize loss''.
The back-propagation algorithm
can be embedded in the \opto optimizer, e.g., an \opto optimizer can use $\tau$ to compute the propagated gradient at each parameter, and apply a gradient descent update. %
\end{example}
\begin{example}[RL]
The parameters are the policy. $g$ is the trajectory (of states, actions, rewards) resulting from running the policy in a Markov decision process; i.e., $g$ documents the graphical model of how a generated action is applied to the transition, which then returns the observation and reward. $f$ can be a success flag. $\omega$ can be ``Maximize return (cumulative rewards)'' or ``Maximize success''.
\end{example}
\begin{example}[Prompt Optimization of an LLM Agent]
The parameters are the prompt of an LLM workflow. $g$ is the computational graph of the agent and $f$ is the feedback about the agent's behavior (which can be scores or natural language). $\omega$ can be ``Maximize score'' or ``Follow the feedback''.
\end{example}

\vspace{-2mm}
\section{\trace: The Next AutoDiff}
\label{sec:design}
\vspace{-1mm}

We design a framework, \trace, to bring \opto from an abstract concept to reality. \trace provides a lightweight Python\footnote{\rev{%
The current implementation is in Python. The DAG-based design of \trace can be adapted to other programming languages also, just like AutoDiff libraries that are available in C++.}} tool to implement the Trace Oracle of \opto for optimizing computational workflows.
Through the \opto framing, \trace separates the design of optimizers and domain-specific components so that general-purpose optimizers can be built that work across diverse domains. %

\vspace{-1mm}
\subsection{Design of \trace} \label{sec:trace core}
\vspace{-1mm}

\trace is designed based on two primitives:
\vspace{-1mm}
\begin{itemize}[leftmargin=*]
    \item \node is the wrapper of Python objects. When wrapped, a Python object is registered as a unique node in the computational graph of \trace. A node can be set \trainable, which would make the node a parameter in \opto. In addition, when using \node to declare a parameter, one can also describe (in natural language) constraints that the parameter should obey.

    \item \bundle is the decorator to turn Python methods into operators. %
    When a function is decorated, its docstring and source code are recorded as the definition of the operator, which infer how the output feedback should change the parameters. %
    Moreover, functions decorated by \bundle can be set \trainable as well, which means that the code of the decorated method becomes a parameter.\footnote{This would add an extra parent (i.e., the trainable code) to the computed child node.}
\end{itemize}

For any workflow, using \trace involves the following steps (see \cref{fig:overall_pipeline}). First, the user declares the workflow's parameters using \node and \bundle, and defines the workflow's conceptual blocks as operators using \bundle.
Then the user creates an \opto optimizer (such as \algo in~\cref{sec:optimizer}), and optionally provides the context $\omega$ for the problem. (A default context $\omega$ of \algo is ``Follow the feedback''). In addition, the user defines a mechanism to provide feedback to the computed result (e.g., scores, natural language suggestions, etc.), in analogy to defining a loss function in neural network training.
After the declaration, optimization via \trace repeats the following:
\begin{enumerate*}[label = \textit{\arabic*)}]
    \item Execute the decorated workflow. As it runs, a DAG is built in the backend, logging the computed results and their connections.
    \item Initiate the propagation of the output feedback to the parameters by calling \backward. (Any execution error is also treated as feedback; see \cref{sec:trace error}.) Internally, \trace extracts the minimal subgraph $g$ connecting the parameters and the output and sends the \opto optimizer the trace feedback $\tau = (f,g)$.
    \item Call the \opto optimizer's \step method to update the parameters.
\end{enumerate*}

\vspace{-1mm}
\subsection{Using \trace Primitives for Effective Execution Tracing}%
\label{sec:constructing opto problem}
\vspace{-1mm}

There are many ways to represent a computational workflow as a computational graph, from abstracting the entire process as one big operator to listing all low-level steps as operators in the graph.
In \trace, the level of abstraction is decided by how \bundle is applied, as all steps underneath \bundle are abstracted as one operator. %
The design of \bundle allows tracing most Python codes, except for those modifying the content of an object reference in place. %
However, such a case can be avoided by first duplicating the object and then applying the modification to the copied object, similar to how a recurrent neural network is implemented. %

Different abstraction choices trade off the graph complexity and the description needed for each operator. Abstracting everything into a single operator makes a trivial graph but requires more descriptions to faithfully capture the workflow.
On the other hand, not all details matter in optimization, so exposing every low-level operator can make the graph unnecessarily cluttered. The best representation depends on the application and \opto optimizer at hand. This problem, we believe, is similar to the design of neural network architectures.
Here, we suggest defining the operators to mimic the whiteboard system diagram of the computational workflow.
This level of abstraction in our experiments strikes a good balance between the ease of documentation and the graph size.

Apart from architecture design, another design question is what information goes into the context $\omega$ versus the description of each operator? For a \emph{single} problem, there is no difference in principle; one can choose to provide details of all operators in $g$ through the context $\omega$. %
However, this will require manually crafting a context for every workflow. We suggest instead providing a description of the operators when they are defined using \bundle. Then \trace will automatically generate the workflow-specific information while the same context $\omega$ is shared across \emph{many} workflows. %

\vspace{-1mm}
\subsection{Backward Feedback Propagation: Realizing the \trace Oracle of OPTO}  \label{sec:backward}
\vspace{-1mm}

\begin{wrapfigure}{o}{0.45\textwidth}
\vspace{-9mm}
\input{algo_boxes}
\vspace{-8mm}
\end{wrapfigure}

\trace uses a recursive graph traversal algorithm (\cref{alg:backward}) to propagate feedback in the reversed topological ordering. %
With different propagators, \cref{alg:backward} can implement various forward-backward schemes %
including back-propagation.\footnote{In back-propagation, the message is the gradient $\nabla_i$ and the \propagate function returns $J_i^\top \sum_j \nabla_j$ to its $i$th parent, where $J_i$ is the Jacobian to the $i$th parent and and the $\nabla_j$ gradient received from the $j$th child.
}  We propose a general propagator, Minimal Subgraph Propagator (\graphpropagator), in \cref{alg:subgraph propagator}. \graphpropagator propagates the trace feedback $\tau=(f,g)$, where $g$ is implemented as a priority queue. %
Running \cref{alg:backward} with \graphpropagator (\cref{alg:subgraph propagator}) together %
implements the Trace Oracle of \opto, which extracts the \emph{minimal subgraph} between parameter nodes and output.\footnote{The \emph{minimal subgraph} $g_{\XX\to Y}$ between nodes $\XX$ and a node $Y$ is defined as $g_{\XX,Y} \coloneqq \XX \bigcup \{ Y\} \bigcup \{ Z, P | Z \in \textrm{ancestors}(Y), Z \in \textrm{descendants}(X), X\in\XX , P \in \textrm{parents}(Z), P \in \textrm{parents}(Y)  \}$.
} connecting the parameters and an output.
\cref{sec:trace analysis} proves the following:

\begin{theorem}
\label{thm:space_complexity}
For a  graph with $N$ nodes and maximum degree $W$, \cref{alg:backward,alg:subgraph propagator} have time complexity $O(W N^2\log N)$ and space complexity\footnote{The space complexity refers to the extra space needed for the backward pass, not including the forward pass.} $O(WN)$.
\end{theorem}

By contrast, back-propagation has time and space complexities of $O(WNd^2)$ and $O(d)$ respectively, where $d$ is the maximal dimension of tensors.
The difference is because in the most general setting of computational graphs and feedback, the propagated feedback (no matter how it is represented) does not have a constant size and needs the full subgraph. %
\begin{theorem}
\label{thm:feedback_length}
For generic computational graphs of $N$ nodes, in the worst case, the propagated feedback needs a description length $\Omega(N)$ to construct an improvement direction.
\end{theorem}

Despite the worst case complexity of \graphpropagator, in practice the difference is negligible.
Since \graphpropagator only involves merging priority queues of references, most actual computation happens in the forward pass (and also the optimizer's \step method).
However for very large problems with millions of nodes in the minimal subgraph, we anticipate that computational issues of \graphpropagator could arise. %

\vspace{-1mm}
\section{Design of the First \opto Optimizer}
\label{sec:optimizer}
\vspace{-1mm}

We introduce an LLM-based generative optimization algorithm \algo for any text-based \opto problem. %
We believe that this is one of many possible optimization algorithms for these problems and there is a large space to be explored for identifying efficient optimization methods for \opto.

\vspace{-1mm}
\paragraph{Subgraph Representation}
One core challenge of designing an LLM-based \opto optimizer is how to represent the execution trace subgraph $g$ (which can involve various graph structures and heterogenous data) to LLMs, in a way that LLMs can understand and reason about the downstream effects of parameter updates.
We leverage the LLMs' remarkable coding and debugging ability~\citep{GithubCopilot}.
\begin{wrapfigure}{h}{0.4\textwidth}
\vspace{-2mm}
\begin{subfigure}{0.4\textwidth}
\begin{lstlisting}[language=Bash,breaklines=true,showstringspaces=false,basicstyle=\fontsize{7}{7}\ttfamily, numbers=none]
#Code:
a = bar(x)
y = add(b, a)
z = mul(a, y)
#Definitions:
[mul] This is a multiply operator.
[add] This is an add operator.
[bar] This is a method that does negative scaling.
#Inputs: 
b=1.0
#Others:
a=2.0    
y=3.0
#Output
z=6.0
#Variable
x=-1.0
#Feedback:
Output should be larger.
\end{lstlisting}
\end{subfigure}
\caption{An example pseudo-code report generated by \trace for a program of \texttt{x = Node(-1.0); z = bar(x) * (bar(x)+1) } and the objective of $\max_x z$.}
\label{fig:code repr}
\vspace{-10mm}
\end{wrapfigure}

We present the trace feedback $(f,g)$ computed by \trace as a pseudo-algorithm problem: the subgraph $g$ is expressed as a report of code execution with information about the computed values and descriptions of operators in $g$. Then we prompt the LLM to update the parameters in $g$ based on feedback $f$ given to the output.
\cref{fig:code repr} shows an example.
It is crucial to note that even though the lines look like an actual program,
it is not the real program but the computational graph defined by \bundle (see \cref{sec:constructing opto problem}).

\vspace{-2mm}
\paragraph{Parameter Update}
We prompt the LLM with a ReAct-CoT style prompt (\cref{sec:optoprime_prompts}) in one query, asking it to generate reasoning based on the graph, %
and a suggestion on the parameter changes. If the suggestion can be extracted from the response, we update the parameters.

\vspace{-2mm}
\paragraph{Optimization Memory}
\algo optimizes most workflows reasonably well using just the traced graph and feedback, but it can run into issues when single feedback alone is not informative (e.g., the output feedback is rewards, but there is no description of how the rewards are generated).
For robustness, we have a basic memory module in \algo, which tracks the past parameter-feedback pairs as in-context examples. See \cref{sec:algo_detail} for details.

\vspace{-1mm}
\section{Experiments}
\label{sec:expts}
\vspace{-1mm}
We evaluate the \trace framework with \algo. %
We implement the state-of-the-art LLM optimizer OPRO~\citep{yang2024large} as a baseline; in comparison with \algo, OPRO does not use the execution trace but relies on the memory of parameter and feedback pairs. For these experiments, we use GPT-4-0125-Preview. We run the experiments on a standard PC with $16$ GB RAM, and \trace introduces no measurable overhead on executing the workflow.  \rev{We also conduct experiments to compare \trace and \algo with a concurrent AutoDiff-like framework, TextGrad~\citep{yuksekgonul2024textgrad}, which was released after \trace was submitted to NeurIPS. We show  that TextGrad can be easily implemented as an optimizer in Trace, and \algo achieves similar or better performance than TextGrad while using much less computation time.}
In the rest of this section, we will denote Trace+\algo simply as \trace. We report the token usages of all approaches in all experiments in \cref{sec:token counts}.

\vspace{-1mm}
\subsection{Validating with Numerical Optimization} \label{sec:numerical exp}
\vspace{-1mm}

First, we want to validate if \algo can solve classical differentiable optimization problems, since they are a special case of \opto. Consider the problem of $\min_x | h(x) - y^*|$ for a target $y^*$.
We construct a synthetic task environment that randomly creates $y^*$ and the computational graph of $h$ with arbitrarily complex connections between numerical variables (see Appendix~\ref{app:sec:numer-opt} for details).
We evaluate \algo (denoted as Trace) and a variant that does not see the graph (Trace Masked); both the optimizers do not use memory. The output feedback is ``The output should be <larger/smaller>'' (this feedback has the same information as the gradient w.r.t. $h$). %
We compare also the performance of Adam optimizer~\citep{kingma2014adam}. We run 30 trials over different randomly generated problems. All methods see the same randomness.
Trace is able to match the best-in-class Adam; on the other hand, without access to the full computational graph, the feedback-alone optimizer struggles to find $x^*$ (Figure~\ref{fig:numer-opt}).

\vspace{-1mm}
\subsection{Tuning Hyperparameters to Orchestrate Complex Systems}
\begin{figure*}
\centering
\begin{subfigure}[t]{0.28\textwidth}
\includegraphics[width=\textwidth]{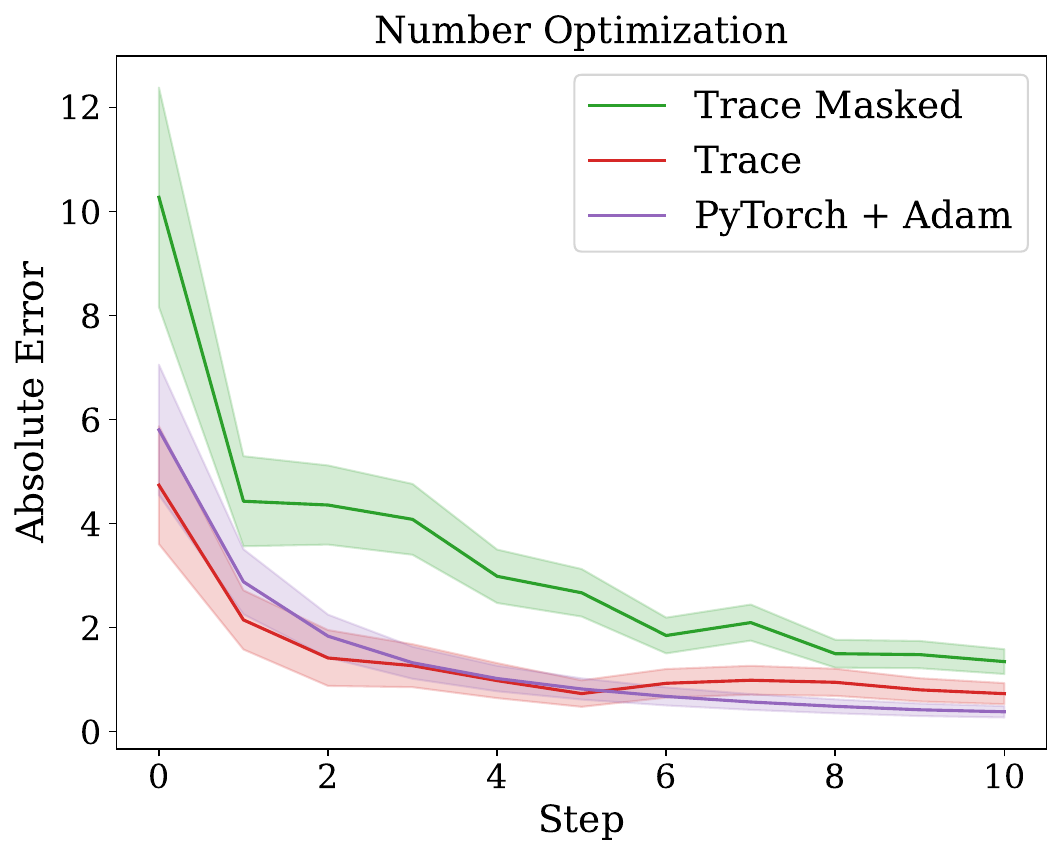}
\caption{Numerical Optimization.}
\label{fig:numer-opt}
\end{subfigure}
~
\begin{subfigure}[t]{0.28\textwidth}
\includegraphics[width=\textwidth]{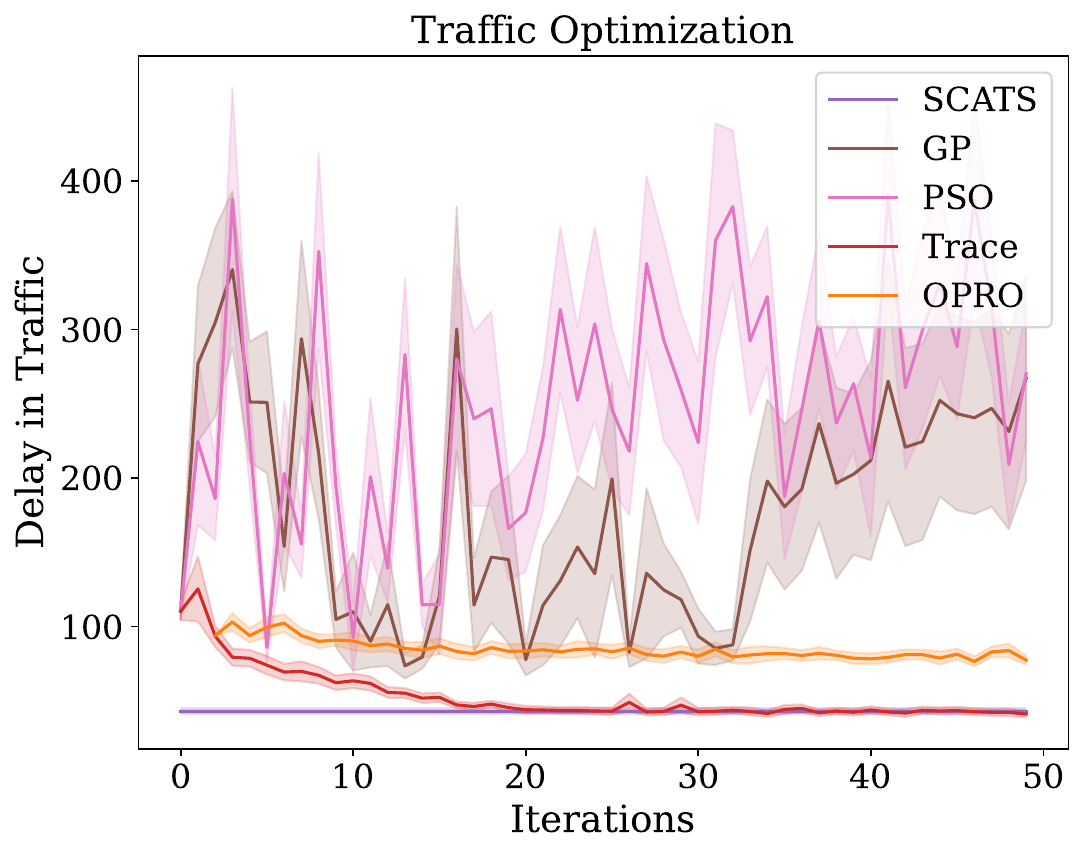}
\caption{\trace vs. Other optimizers.}
\end{subfigure}
~
\begin{subfigure}[t]{0.28\textwidth}
\includegraphics[width=\textwidth]{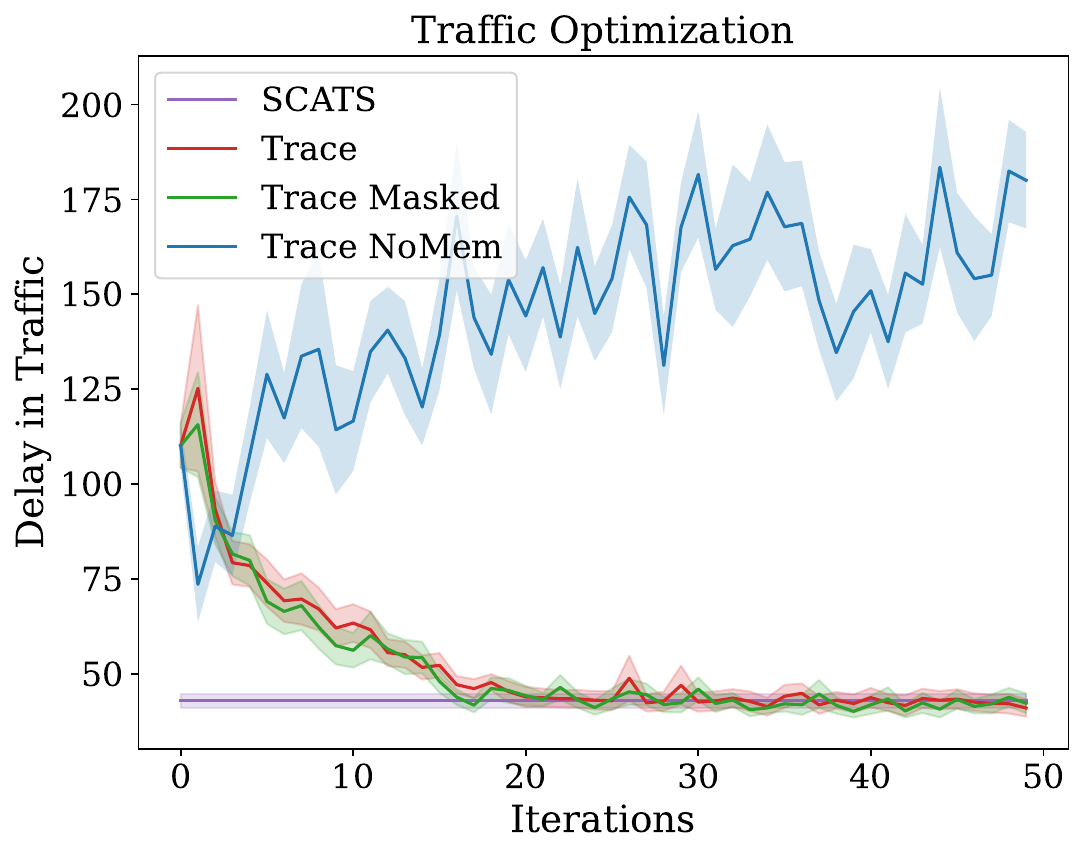}
\caption{Ablations of \trace.}
\end{subfigure}
\caption{Numerical Optimization and Traffic Optimization Results.}
\label{fig:Traffic_expts}
\vspace{-4mm}
\end{figure*}

\vspace{-1mm}
We tested \trace in a traffic control problem, which is an instance of hyper-parameter tuning. We used UXSim~\citep{seo2023uxsim} to simulate traffic at a four-way intersection, where the trainable parameters are 2 integers in $[15,90]$, which are the green light duration for each direction of traffic flow.
The feedback is %
the estimated delay experienced by all vehicles due to intersections, and the goal of an optimizer is to %
minimize the delay using the fewest number of traffic simulations. To this end, this optimizer must find the right trade-off for temporally distributed and variable demands.
In \cref{fig:Traffic_expts} we report the performance of a SOTA heuristic from the traffic control literature, SCATS~\citep{scats} %
as well as two black-box optimization techniques: Gaussian Process Minimization (GP)~\citep{frazier2018bayesian} and Particle Swarm Optimization (PSO)~\citep{kennedy1995particle}. All methods use the same starting parameters. %
\rev{Trace denotes \algo using memory, and Trace NoMem denotes \algo without memory.}
We report further details in Appendix~\ref{app:sec:traffic}.
GP and PSO appear bad because $50$ iterations are insufficient for their convergence; given enough iterations, both will eventually perform well.
\trace is quickly competitive with the SCATS heuristic, whereas OPRO is not. Moreover, we find that memory is crucial for \trace to perform well for this task. But we note that \trace consumes extra overhead compared to other methods, since \trace has to materialize the resulting computation graph and query an LLM with effectively a longer prompt than that of OPRO.

\subsection{Unifying Prompts and Functions Optimization}

Many LLM agents today, e.g., specified by LangChain~\citep{langchain-blog} and DSPy~\citep{khattab2023dspy}, have many components.
These libraries provide optimization tools to optimize a small portion of their workflows, predominantly the prompt that goes into an LLM call. %
However, for building self-adapting agents that can modify their own behavior, only allowing the change to one part of a workflow but not others can be limiting. %
In this experiment, we test \trace's ability in joint prompt optimization and code generation.
Specifically, we optimize a given DSPy-based LLM agent and tune its three components: the meta-prompt \verb|prompt_template|, a function \verb|create_prompt| that modifies the prompt with the current question, and a function \verb|extract_answer| that post-processes the output of an LLM call. %

We set up an end-to-end prompt-and-code optimization pipeline.
We use an automatic evaluation function to compare the LLM's output with the ground truth, which requires the LLM agent to generate outputs not only with the correct answer but also in the correct format \rev{(following the guidelines of~\citep{liu2023agentbench})}. %
We use the Big-Bench Hard dataset~\citep{suzgun2022challenging} %
(15 examples for training, 5 for validation, and the rest for testing).
We compare Trace with DSPy's COPRO module (which optimizes the meta-prompt).
In Table~\ref{tab:bbh-hard}, we show that Trace is able to optimize a DSPy program beyond what DSPy's COPRO optimizer can, especially on algorithmic tasks. This result shows how \trace can concretely improve upon existing LLM prompt optimization libraries. We show learned codes in \cref{sec:optimized params examples}.

\begin{table}[htb]
\footnotesize
\centering
\resizebox{\columnwidth}{!}{
\begin{tabular}{@{}rccc|rccc@{}}
\toprule
        & BBH all       & NLP           & Algorithmic   & \multicolumn{1}{c}{} & BBH all       & NLP           & Algorithmic   \\
    0-shot    & (23 tasks)    & (12 tasks)    & (11 tasks)    & \multicolumn{1}{c}{0-shot} & (23 tasks)    & (12 tasks)    & (11 tasks)    \\ \midrule
DSPy    & 41.6          & 53.8          & 32.6          & DSPy + CoT             & 70.4          & 73.7          & 68.0          \\
DSPy-PO & 55.3          & 69.0          & 45.2          & DSPy-PO + CoT          & 71.6          & 73.9          & 70.0          \\ \midrule
Trace   & \textbf{59.5} & \textbf{70.9} & \textbf{51.1} & Trace + CoT            & \textbf{78.6} & \textbf{75.8} & \textbf{80.6} \\ \bottomrule
\end{tabular}
}
\vspace{1mm}
\caption{End-to-end workflow optimization for an LLM benchmark (Big-Bench Hard) in 0-shot setup. \texttt{CoT} refers to Chain-of-Thought prompting and \texttt{PO} refers to DSPy's own prompt optimizer (COPRO). We use Trace to optimize a DSPy program, starting from the same program and prompt template specified by DSPy.}
\label{tab:bbh-hard}
\vspace{-4mm}
\end{table}

\vspace{-1mm}
\subsection{Long-Horizon Robot Manipulator Control}
\vspace{-1mm}

We test the ability of \trace to optimize long-horizon workflows with complex dependencies and to ``back-propagate through time''.
We use \trace to train the controller code (in Python) for a simulated Sawyer robot manipulator.
We use the Meta-World environment from LLF-Bench~\citep{cheng2023llf} as the simulator and consider three tasks: Reach, Pick-place, and Push. For each task, LLF-Bench provides a task instruction and meaning of the action space, which we use as the context $\omega$ of the \opto problem.
The observation is a Python dictionary of vectors, indicating the end-effector position, the goal position, the gripper status, etc. The action space is a 4-dimensional vector to control the relative position of the end-effector and the gripper state. In each time step, the LLF-Bench Meta-World simulator returns the observation and natural language feedback to guide the robot. An episode ends if the robot successfully solves the problem or because of time-out.
We consider an episodic training setting. The initial condition for all iterations in training is the same. %
We evaluate the learned policy in terms of success, starting from 10 held-out initial conditions. The task horizon is 10 steps, which is sufficient for task completion, and each training iteration has one rollout. The output feedback in \opto is a textual representation of task success. %
In addition to the controller code, we also decorated the reset and step functions of the gym environment so that the entire rollout can be traced end-to-end. %
We compare \trace with OPRO; because of the streaming \opto setting, our OPRO implementation only proposes one candidate in each iteration, which is then evaluated and provided with the output feedback.

\begin{figure*}
\centering
\begin{subfigure}[t]{0.16\textwidth}
\centering
\includegraphics[trim={10.8cm 4cm 10.8cm 13cm},clip,width=\textwidth]{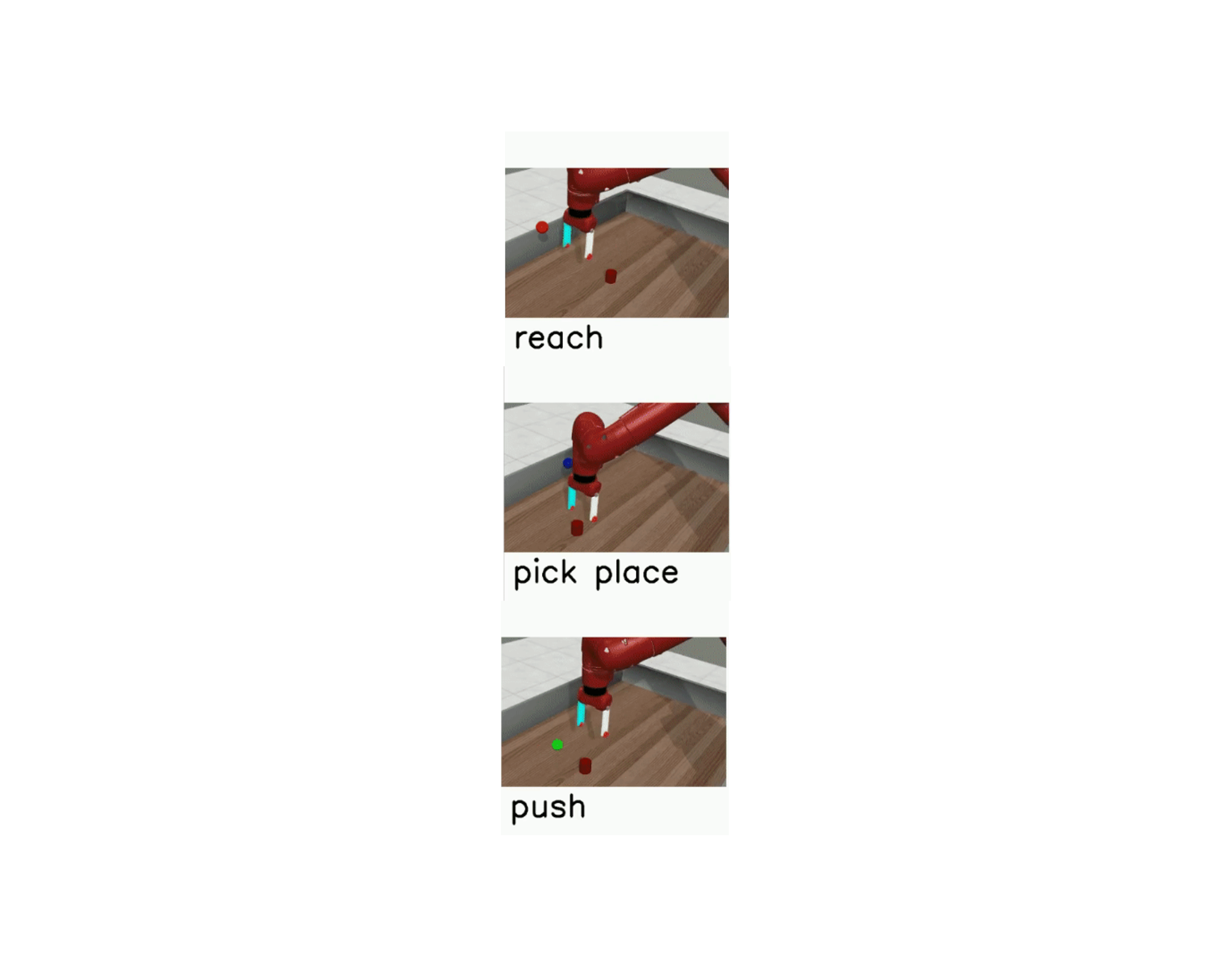}
\end{subfigure}
\begin{subfigure}[t]{0.80\textwidth}
\centering
\begin{subfigure}[t]{0.32\textwidth}
\includegraphics[width=0.96\textwidth]{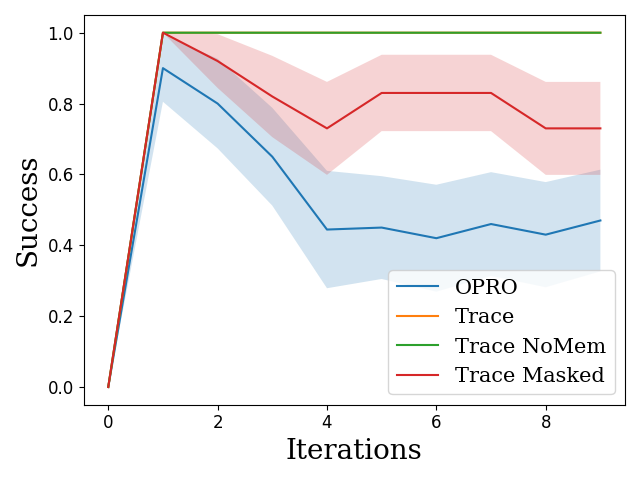}
\caption{Reach (Test)}
\end{subfigure}
\begin{subfigure}[t]{0.32\textwidth}
\includegraphics[width=0.96\textwidth]{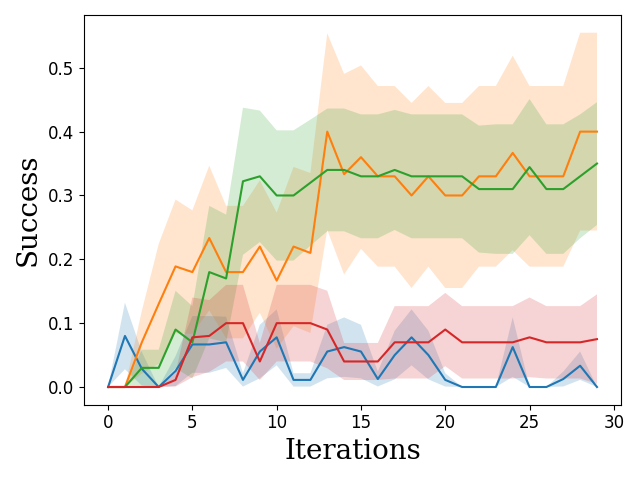}
\caption{Pick-place (Test)}
\end{subfigure}
\begin{subfigure}[t]{0.32\textwidth}
\includegraphics[width=0.96\textwidth]{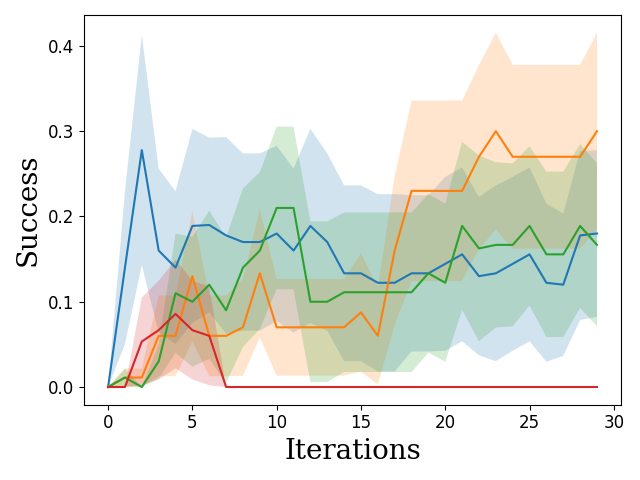}
\caption{Push (Test)}
\end{subfigure}
\end{subfigure}

\caption{Learning the feedback control policy (code) for a simulated Sawyer manipulator in LLF-Bench Meta-World. In each iteration (x-axis), one episode of rollout (10 steps) is performed, and then the policy is updated. The mean and standard error of the success rate over 10 seeds are shown.}
\label{fig:mw_exps}
\vspace{-4mm}
\end{figure*}

The experimental results are summarized in \cref{fig:mw_exps}.
\rev{Trace denotes \algo using memory, and Trace NoMem denotes \algo without memory.}
We show learned code in \cref{sec:optimized params examples}. %
\algo is clearly the top-performing optimizer, especially the version with memory. OPRO is able to solve Reach at the start, but its performance degraded over iterations (this instability was mentioned in \citep{yang2024large}) and gets a similar performance as \algo (without memory) in Push. To validate that the performance of \algo is indeed due to using the execution trace, we include an ablation where we mask out the execution trace, which leads to a significant decline in performance and stability.
This experiment features the most complex graph structures among all the experiments. %
The experimental results here are quite impressive, showing that \trace is able to learn a sophisticated control logic in dozens of interactions, not only working on the training initial conditions but also on held-out testing ones too. We discuss some limitations in Appendix~\ref{app:sec:metaworld}.

\vspace{-1mm}
\subsection{Comparison with TextGrad} \label{sec:exp of textgrad}
\vspace{-1mm}

\rev{
After the submission of our work to NeurIPS, another AutoDiff-like framework, TextGrad~\citep{yuksekgonul2024textgrad}, was released, which shares the same goal of end-to-end optimizing AI workflows as \trace.  In comparison, TextGrad propagates text feedback, whereas \trace propagates minimal subgraphs (see \cref{sec:backward}). The graph-based design of \trace, which separates the tracing infrastructure and optimization algorithms, makes it more flexible.
In fact, we easily implemented TextGrad as an optimizer in the \trace framework, but the reverse is not possible (because TextGrad couples the infrastructure and the optimization algorithm together). In addition, unlike TextGrad, \trace supports jointly optimizing heterogeneous parameters and can be applied to directly trace a given computational workflow without the need to rewrite the workflow using pre-defined templates.
Please see~\cref{sec:textgrad_comparison} for more discussion comparing the two frameworks.
}

\rev{
In this experiment,  we apply \trace to directly decorate the evaluation code released with the TextGrad library and optimize the parameters following their training/evaluation pipeline line-by-line. This experimental design makes the comparison fair by allowing each optimizer to access the same LLM APIs around the same time, and showcases the flexibility of \trace framework to optimize any computational workflow.
We pick the Solution optimization~\citep[Table 2]{yuksekgonul2024textgrad} and Prompt optimization~\citep[Table 3]{yuksekgonul2024textgrad} for the reasoning tasks experiments. Please see \citep{yuksekgonul2024textgrad} for details on the exact setup. We compare \algo, TextGrad\footnote{The numbers in the original paper cannot be reproduced exactly despite using the released TextGrad code.}, and a reimplementation\footnote{All algorithms in Trace see the same graph, which is slightly different from the graph of TextGrad.} of TextGrad as an optimizer in \trace. We find that all these algorithms achieve similar success rates in these experiments. One noticeable difference is that \algo is about $3$x faster wall-clock time than TextGrad since \algo makes a single call to LLM in each optimization step, whereas TextGrad calls linear to the graph's size.
}

\begin{table}[ht]
\footnotesize
\centering
\resizebox{\columnwidth}{!}{
    \begin{tabular}{@{}c|c|c|c|c|c|c@{}}
    \toprule
          & \begin{tabular}[c]{@{}c@{}}OptoPrime \\ (Trace)\end{tabular} & Time & \begin{tabular}[c]{@{}c@{}}TextGrad \\ (24-10-30)\end{tabular} & Time & \begin{tabular}[c]{@{}c@{}}TextGrad \\ (Trace)\end{tabular} & \begin{tabular}[c]{@{}c@{}}TextGrad \\ (Reported)\end{tabular}  \\
         \midrule
         MMLU-Machine Learning  & \textbf{86.6} (0.2)  & 1.7 (0.6) & 86.1 (0.5) & 3.5 (1.1)  & 86.3 (0.2) & \gray{88.4} \\
        MMLU-College Physics  &  \textbf{94.1} (0.8) & 1.2 (0.3) & 93.1 (0.7)  & 2.3 (0.4) & 93.3 (0.6) & \gray{95.1} \\
        Google-proof QA & \textbf{59.6} (1.3)  & 12.2 (1.4)  & 53.2 (0.6) & 19.5 (1.9) & 54.0 (0.7) & \gray{55.0} \\ \midrule
        BBH Counting  &   \textbf{89.4} (0.1) & 55.9 (4.5) & 89.2 (1.2) & 142.9 (9.3) & 87.6 (1.7) & \gray{91.9} \\

        BBH Word Sorting  &   71.6 (3.1) & 82.5 (10.1) &  \textbf{72.0} (0.4) & 211.1 (16.8) & 71.4 (2.5) & \gray{79.8} \\
        GSM8K &  \textbf{82.5} (0.1) & --- & 82.4 (0.6)  & ---  &  82.0 (0.2) & \gray{81.1} \\
 \bottomrule
    \end{tabular}}
    \vspace{1mm}
\caption{\textbf{Comparison between \trace and TextGrad.} The optimizer is GPT-4o-2024-08-06, and the student model is GPT-35-turbo-1106. The results show the mean and the standard error of success rate of the last iterate computed by 5 seeds. The experiment time reported is in minutes \rev{(the time involves not just training but also validation and testing by running TextGrad's original pipeline)}; the time of GSM8K experiment is omitted as the experiment time (>8hrs) is determined primarily by the evaluation not optimization. }
    \label{fig:textgrad exp}
    \vspace{-4mm}
\end{table}

\vspace{-1mm}
\section{Limitations}
\label{sec:limitations}
\vspace{-1mm}

We highlight that \trace, \opto and \algo are a first step towards end-to-end generative optimization and building self-adapting workflows. They have limitations in their current form. 
\opto captures rich feedback, but it is important to specify a solution concept as well as the feedback source. We provide guidance for feedback design in~\cref{sec:constructing opto problem} and discuss notions of optimality in~\cref{sec:opto solution concept}. \rev{We believe designing feedback will be as important as designing loss function in deep learning, both of which are open research questions.}
Also, \trace cannot convert all computational workflows into \opto problems, e.g., stateful functions that modify their state in place cannot be represented as a DAG without modification, %
and distributed/parallel computing workflows are incompatible with the current implementation \rev{(though in theory \trace can run in an asynchronous way so long as the overall graph does not end up with cycles). } %
\rev{Finally, while \trace is designed to be generic and future-proof, the %
\algo optimizer is preliminary.}
Although we demonstrated that \algo could work well with moderate-size graphs, it is not a provably optimal algorithm \rev{and uses more tokens than OPRO, though, in our experiments, OPRO's performance does not improve even when given a large token budget.}
The debugging ability and context limits of the LLM used in \algo crucially determine the scale of problems that we can practically address today. %
\rev{Consequently, more research is needed for designing token-efficient generative optimization algorithms.}

\vspace{-1mm}
\section{Conclusion and Future Work}
\label{sec:conclusion}
\vspace{-1mm}

We created \trace that can convert a computational workflow optimization problem into an \opto problem, and we 
demonstrated a tractable \opto optimizer, \algo.
This is just a first step towards a new paradigm of end-to-end \emph{generative optimization}, with exciting avenues for future work. We discuss a few selected ones below. Please see \cref{sec:Deep Agent Workflows} for a longer discussion. 

In \algo, we connect optimization to an LLM's reasoning capability. 
Techniques that have been proposed to improve LLM reasoning, e.g. Chain-of-Thought~\citep{wei2022chain}, Few-Shot Prompting~\citep{fewshot}, Tool Use~\citep{schick2023toolformer}, and Multi-Agent Workflows~\citep{wu2023autogen} could also help improve \algo or design new \opto optimizers. 
We conjecture that a hybrid workflow of LLM and search algorithms, %
can enable a truly general-purpose \opto optimizer. Along the way, we must settle on how to delineate the agent vs. the optimizer. How to trade off the generality of optimizer vs. crafting side-information in the context $\omega$ to achieve task-specific performance is an open question.

In \trace, we chose a specific propagator (\graphpropagator), which maximally preserves information for a general computation graph. 
We can instead specialize it for specific computations, e.g. to accommodate very large graphs with a hierarchical graph representation.
Going a step beyond the basic memory module we experimented with in \algo, we anticipate that an optimizer that can reason about how a workflow will behave under counterfactual parameter settings %
can be more efficient than \algo and can enable a divide-and-conquer approach to \opto. 
\rev{More research is needed to study the theoretical properties of \opto (such as optimization landscape and complexity). We hope our preliminary effort in \cref{sec:opto solution concept} can provide some guidance.}
Finally, in this paper, we focused on output feedback and context that can be compactly textualized. We anticipate that computational workflows with rich non-textual contexts and output feedback will also benefit from automatic generative optimization through appropriate applications of \trace (e.g., with VLMs).

\vspace{-2mm}
\begin{ack}
We would like to thank John Langford, Ahmed Awadallah, Jennifer Neville, Andrey Kolobov, Ricky Loynd and Paul Mineiro for thought-provoking discussions.
We would also like to thank Tobias Schnabel, Ruijie Zheng, Wanqiao Xu, and Kaiwen Wang for their valuable feedback on an early draft of this manuscript.
Additionally, we thank Anqi Li, Omar Khattab, David Hall, Yifan Mai, Bryan He, Yash Chandak, Emma Brunksill, and Dawen Liang for their suggestions and feedback. The work was partially done while Adith Swaminathan was at Microsoft Research.
\end{ack}

\bibliographystyle{unsrt}
\bibliography{references}

\begin{thebibliography}{10}

\bibitem{compound-ai-blog}
Matei Zaharia, Omar Khattab, Lingjiao Chen, Jared~Quincy Davis, Heather Miller,
  Chris Potts, James Zou, Michael Carbin, Jonathan Frankle, Naveen Rao, and Ali
  Ghodsi.
\newblock The shift from models to compound {AI} systems.
\newblock \url{https://bair.berkeley.edu/blog/2024/02/18/compound-ai-systems/},
  2024.

\bibitem{openai2023chatgpt}
Josh Achiam, Steven Adler, Sandhini Agarwal, Lama Ahmad, Ilge Akkaya,
  Florencia~Leoni Aleman, Diogo Almeida, Janko Altenschmidt, Sam Altman,
  Shyamal Anadkat, et~al.
\newblock {GPT}-4 technical report.
\newblock {\em arXiv preprint arXiv:2303.08774}, 2023.

\bibitem{GithubCopilot}
Albert Ziegler, Eirini Kalliamvakou, X.~Alice Li, Andrew Rice, Devon Rifkin,
  Shawn Simister, Ganesh Sittampalam, and Edward Aftandilian.
\newblock Productivity assessment of neural code completion.
\newblock In {\em SIGPLAN International Symposium on Machine Programming}, page
  21–29, 2022.

\bibitem{liang2023code}
Jacky Liang, Wenlong Huang, Fei Xia, Peng Xu, Karol Hausman, Brian Ichter, Pete
  Florence, and Andy Zeng.
\newblock Code as policies: Language model programs for embodied control.
\newblock In {\em IEEE International Conference on Robotics and Automation},
  pages 9493--9500, 2023.

\bibitem{wu2023autogen}
Qingyun Wu, Gagan Bansal, Jieyu Zhang, Yiran Wu, Shaokun Zhang, Erkang Zhu,
  Beibin Li, Li~Jiang, Xiaoyun Zhang, and Chi Wang.
\newblock Autogen: Enabling next-gen {LLM} applications via multi-agent
  conversation framework.
\newblock {\em arXiv preprint arXiv:2308.08155}, 2023.

\bibitem{conn2009introduction}
Andrew~R Conn, Katya Scheinberg, and Luis~N Vicente.
\newblock {\em Introduction to derivative-free optimization}.
\newblock SIAM, 2009.

\bibitem{hutter2011sequential}
Frank Hutter, Holger~H Hoos, and Kevin Leyton-Brown.
\newblock Sequential model-based optimization for general algorithm
  configuration.
\newblock In {\em Learning and Intelligent Optimization}, pages 507--523.
  Springer, 2011.

\bibitem{frazier2018bayesian}
Peter~I Frazier.
\newblock Bayesian optimization.
\newblock {\em Recent Advances in Optimization and Modeling of Contemporary
  Problems}, pages 255--278, 2018.

\bibitem{back1993overview}
Thomas B{\"a}ck and Hans-Paul Schwefel.
\newblock An overview of evolutionary algorithms for parameter optimization.
\newblock {\em Evolutionary computation}, 1(1):1--23, 1993.

\bibitem{ginebra1995response}
Josep Ginebra and Murray~K Clayton.
\newblock Response surface bandits.
\newblock {\em Journal of the Royal Statistical Society Series B: Statistical
  Methodology}, 57(4):771--784, 1995.

\bibitem{zhou2022large}
Yongchao Zhou, Andrei~Ioan Muresanu, Ziwen Han, Keiran Paster, Silviu Pitis,
  Harris Chan, and Jimmy Ba.
\newblock Large language models are human-level prompt engineers.
\newblock In {\em ICLR}, 2023.

\bibitem{pryzant2023automatic}
Reid Pryzant, Dan Iter, Jerry Li, Yin Lee, Chenguang Zhu, and Michael Zeng.
\newblock Automatic prompt optimization with “gradient descent” and beam
  search.
\newblock In {\em EMNLP}, pages 7957--7968, 2023.

\bibitem{yang2024large}
Chengrun Yang, Xuezhi Wang, Yifeng Lu, Hanxiao Liu, Quoc~V Le, Denny Zhou, and
  Xinyun Chen.
\newblock Large language models as optimizers.
\newblock In {\em ICLR}, 2024.

\bibitem{schnabel2024prompts}
Tobias Schnabel and Jennifer Neville.
\newblock Prompts as programs: A structure-aware approach to efficient
  compile-time prompt optimization.
\newblock {\em arXiv preprint arXiv:2404.02319}, 2024.

\bibitem{zelikman2023self}
Eric Zelikman, Eliana Lorch, Lester Mackey, and Adam~Tauman Kalai.
\newblock Self-taught optimizer ({STOP}): Recursively self-improving code
  generation.
\newblock {\em arXiv preprint arXiv:2310.02304}, 2023.

\bibitem{nie2023importance}
Allen Nie, Ching-An Cheng, Andrey Kolobov, and Adith Swaminathan.
\newblock The importance of directional feedback for {LLM}-based optimizers.
\newblock {\em arXiv preprint arXiv:2405.16434}, 2024.

\bibitem{rumelhart1986learning}
David~E Rumelhart, Geoffrey~E Hinton, and Ronald~J Williams.
\newblock Learning representations by back-propagating errors.
\newblock {\em Nature}, 323(6088):533--536, 1986.

\bibitem{AutoDiff}
Atilim~Gunes Baydin, Barak~A. Pearlmutter, Alexey~Andreyevich Radul, and
  Jeffrey~Mark Siskind.
\newblock Automatic differentiation in machine learning: a survey.
\newblock {\em Journal of Machine Learning Research}, 18(153):1--43, 2018.

\bibitem{paszke2019pytorch}
Adam Paszke, Sam Gross, Francisco Massa, Adam Lerer, James Bradbury, Gregory
  Chanan, Trevor Killeen, Zeming Lin, Natalia Gimelshein, Luca Antiga, et~al.
\newblock {PyTorch}: an imperative style, high-performance deep learning
  library.
\newblock In {\em NeurIPS}, pages 8026--8037, 2019.

\bibitem{nagaraj2012structured}
Karthik Nagaraj, Charles Killian, and Jennifer Neville.
\newblock Structured comparative analysis of systems logs to diagnose
  performance problems.
\newblock In {\em USENIX Symposium on Networked Systems Design and
  Implementation}, pages 353--366, 2012.

\bibitem{suzgun2022challenging}
Mirac Suzgun, Nathan Scales, Nathanael Sch{\"a}rli, Sebastian Gehrmann, Yi~Tay,
  Hyung~Won Chung, Aakanksha Chowdhery, Quoc Le, Ed~Chi, Denny Zhou, et~al.
\newblock Challenging {BIG-Bench} tasks and whether chain-of-thought can solve
  them.
\newblock In {\em Findings of the ACL}, pages 13003--13051, 2023.

\bibitem{khattab2023dspy}
Omar Khattab, Arnav Singhvi, Paridhi Maheshwari, Zhiyuan Zhang, Keshav
  Santhanam, Sri Vardhamanan, Saiful Haq, Ashutosh Sharma, Thomas~T Joshi,
  Hanna Moazam, et~al.
\newblock {DSPy}: Compiling declarative language model calls into
  self-improving pipelines.
\newblock {\em arXiv preprint arXiv:2310.03714}, 2023.

\bibitem{cheng2023llf}
Ching-An Cheng, Andrey Kolobov, Dipendra Misra, Allen Nie, and Adith
  Swaminathan.
\newblock {LLF-Bench}: Benchmark for interactive learning from language
  feedback.
\newblock {\em arXiv preprint arXiv:2312.06853}, 2023.

\bibitem{yuksekgonul2024textgrad}
Mert Yuksekgonul, Federico Bianchi, Joseph Boen, Sheng Liu, Zhi Huang, Carlos
  Guestrin, and James Zou.
\newblock {TextGrad}: Automatic ``differentiation'' via text.
\newblock {\em arXiv preprint arXiv:2406.07496}, 2024.

\bibitem{kingma2014adam}
Diederik~P Kingma and Jimmy Ba.
\newblock Adam: A method for stochastic optimization.
\newblock {\em arXiv preprint arXiv:1412.6980}, 2014.

\bibitem{seo2023uxsim}
Toru Seo.
\newblock {UXsim}: An open source macroscopic and mesoscopic traffic simulator
  in python--a technical overview.
\newblock {\em arXiv preprint arXiv:2309.17114}, 2023.

\bibitem{scats}
Courtney Slavin, Wei Feng, Miguel Figliozzi, and Peter Koonce.
\newblock Statistical study of the impact of adaptive traffic signal control on
  traffic and transit performance.
\newblock {\em Transportation Research Record}, 2356(1):117--126, 2013.

\bibitem{kennedy1995particle}
James Kennedy and Russell Eberhart.
\newblock Particle swarm optimization.
\newblock In {\em ICNN}, volume~4, pages 1942--1948, 1995.

\bibitem{langchain-blog}
LangChain Team.
\newblock Langchain tracing.
\newblock \url{https://blog.langchain.dev/tracing/}, 2023.

\bibitem{liu2023agentbench}
Xiao Liu, Hao Yu, Hanchen Zhang, Yifan Xu, Xuanyu Lei, Hanyu Lai, Yu~Gu,
  Hangliang Ding, Kaiwen Men, Kejuan Yang, Shudan Zhang, Xiang Deng, Aohan
  Zeng, Zhengxiao Du, Chenhui Zhang, Sheng Shen, Tianjun Zhang, Yu~Su, Huan
  Sun, Minlie Huang, Yuxiao Dong, and Jie Tang.
\newblock {AgentBench}: Evaluating {LLM}s as agents.
\newblock In {\em ICLR}, 2024.

\bibitem{wei2022chain}
Jason Wei, Xuezhi Wang, Dale Schuurmans, Maarten Bosma, Brian Ichter, Fei Xia,
  Ed~H Chi, Quoc~V Le, and Denny Zhou.
\newblock Chain-of-thought prompting elicits reasoning in large language
  models.
\newblock In {\em NeurIPS}, pages 24824--24837, 2022.

\bibitem{fewshot}
Tom Brown, Benjamin Mann, Nick Ryder, Melanie Subbiah, Jared~D Kaplan, Prafulla
  Dhariwal, Arvind Neelakantan, Pranav Shyam, Girish Sastry, Amanda Askell,
  et~al.
\newblock Language models are few-shot learners.
\newblock In {\em NeurIPS}, 2020.

\bibitem{schick2023toolformer}
Timo Schick, Jane Dwivedi-Yu, Roberto Dessi, Roberta Raileanu, Maria Lomeli,
  Eric Hambro, Luke Zettlemoyer, Nicola Cancedda, and Thomas Scialom.
\newblock Toolformer: Language models can teach themselves to use tools.
\newblock In {\em NeurIPS}, 2023.

\bibitem{bwgames}
James Wexler.
\newblock {Artificial Intelligence in Games: A look at the smarts behind
  Lionhead Studio’s ``Black and White'' and where it can and will go in the
  future}.
\newblock
  \url{https://www.cs.rochester.edu/~brown/242/assts/termprojs/games.pdf},
  2002.

\bibitem{yao2024retroformer}
Weiran Yao, Shelby Heinecke, Juan~Carlos Niebles, Zhiwei Liu, Yihao Feng,
  Le~Xue, Rithesh~R N, Zeyuan Chen, Jianguo Zhang, Devansh Arpit, Ran Xu,
  Phil~L Mui, Huan Wang, Caiming Xiong, and Silvio Savarese.
\newblock Retroformer: Retrospective large language agents with policy gradient
  optimization.
\newblock In {\em ICLR}, 2024.

\bibitem{sordoni2024joint}
Alessandro Sordoni, Xingdi Yuan, Marc-Alexandre C{\^o}t{\'e}, Matheus Pereira,
  Adam Trischler, Ziang Xiao, Arian Hosseini, Friederike Niedtner, and Nicolas
  Le~Roux.
\newblock Joint prompt optimization of stacked {LLM}s using variational
  inference.
\newblock In {\em NeurIPS}, 2023.

\bibitem{zhuge2024language}
Mingchen Zhuge, Wenyi Wang, Louis Kirsch, Francesco Faccio, Dmitrii Khizbullin,
  and Jurgen Schmidhuber.
\newblock Language agents as optimizable graphs.
\newblock {\em arXiv preprint arXiv:2402.16823}, 2024.

\bibitem{ma2024large}
Ruotian Ma, Xiaolei Wang, Xin Zhou, Jian Li, Nan Du, Tao Gui, Qi~Zhang, and
  Xuanjing Huang.
\newblock Are large language models good prompt optimizers?
\newblock {\em arXiv preprint arXiv:2402.02101}, 2024.

\bibitem{wang2023promptagent}
Xinyuan Wang, Chenxi Li, Zhen Wang, Fan Bai, Haotian Luo, Jiayou Zhang, Nebojsa
  Jojic, Eric Xing, and Zhiting Hu.
\newblock {PromptAgent}: Strategic planning with language models enables
  expert-level prompt optimization.
\newblock In {\em ICLR}, 2024.

\bibitem{tang2024unleashing}
Xinyu Tang, Xiaolei Wang, Wayne~Xin Zhao, Siyuan Lu, Yaliang Li, and Ji-Rong
  Wen.
\newblock Unleashing the potential of large language models as prompt
  optimizers: An analogical analysis with gradient-based model optimizers.
\newblock {\em arXiv preprint arXiv:2402.17564}, 2024.

\bibitem{chen2023teaching}
Xinyun Chen, Maxwell Lin, Nathanael Sch{\"a}rli, and Denny Zhou.
\newblock Teaching large language models to self-debug.
\newblock In {\em ICLR}, 2024.

\bibitem{ma2024llamoco}
Zeyuan Ma, Hongshu Guo, Jiacheng Chen, Guojun Peng, Zhiguang Cao, Yining Ma,
  and Yue-Jiao Gong.
\newblock {LLaMoCo}: Instruction tuning of large language models for
  optimization code generation.
\newblock {\em arXiv preprint arXiv:2403.01131}, 2024.

\bibitem{liu2024large}
Tennison Liu, Nicol{\'a}s Astorga, Nabeel Seedat, and Mihaela van~der Schaar.
\newblock Large language models to enhance bayesian optimization.
\newblock In {\em ICLR}, 2024.

\bibitem{zhang2023using}
Michael~R. Zhang, Nishkrit Desai, Juhan Bae, Jonathan Lorraine, and Jimmy Ba.
\newblock Using large language models for hyperparameter optimization.
\newblock {\em arXiv preprint arXiv:2312.04528}, 2023.

\bibitem{bartok2014partial}
G{\'a}bor Bart{\'o}k, Dean~P Foster, D{\'a}vid P{\'a}l, Alexander Rakhlin, and
  Csaba Szepesv{\'a}ri.
\newblock Partial monitoring—classification, regret bounds, and algorithms.
\newblock {\em Mathematics of Operations Research}, pages 967--997, 2014.

\bibitem{hascoet2006enabling}
Laurent Hascoet and Mauricio Araya-Polo.
\newblock Enabling user-driven checkpointing strategies in reverse-mode
  automatic differentiation.
\newblock {\em arXiv preprint cs/0606042}, 2006.

\bibitem{shaban2019truncated}
Amirreza Shaban, Ching-An Cheng, Nathan Hatch, and Byron Boots.
\newblock Truncated back-propagation for bilevel optimization.
\newblock In {\em AISTATS}, pages 1723--1732, 2019.

\bibitem{cheng2020online}
Ching-An Cheng, Jonathan Lee, Ken Goldberg, and Byron Boots.
\newblock Online learning with continuous variations: Dynamic regret and
  reductions.
\newblock In {\em AISTATS}, pages 2218--2228, 2020.

\bibitem{kakade2003sample}
Sham~Machandranath Kakade.
\newblock {\em On the Sample Complexity of Reinforcement Learning}.
\newblock PhD thesis, University College London, 2003.

\end{thebibliography}

\appendix

\setcounter{figure}{0}
\renewcommand{\thefigure}{A.\arabic{figure}}
\setcounter{table}{0}  %
\renewcommand{\thetable}{A.\arabic{table}}

\newpage

\section{Perspective: Deep Agent Workflows} \label{sec:Deep Agent Workflows}

We posit that the current practice of manually engineering computational workflows to build AI systems is analogous to programmers in the early 2000s hand-coding neural network weights to create engaging AI characters in video games~\citep{bwgames}.  
Just like AutoDiff enabled the automatic and scalable optimization of deep neural networks with billions of parameters, we believe that \trace is the first step towards automatic and scalable optimization of ``Deep Agent Workflows'' to power even more capable AI systems. However, there are several limitations of the current implementation of \trace that need to be addressed to build Deep Agent Workflows.  

When designing interactive AI systems that learn from their interactions, we need to define the \emph{parameters} and \emph{feedback} of the system. Parameters are the internal attributes that can be updated by the learning algorithm employed by the system. Feedback are the things observed and recorded by the system as a product of its interactions, and that provide signal for learning. 
\trace enables the development of new learning algorithms (e.g. through \algo) that incorporate rich feedback to update heterogenous parameters. In contrast, AutoDiff for deep neural networks uses numerical feedback (e.g. rewards or loss functions) to optimize numerical parameters (e.g. tensors). Black-box optimization techniques (e.g. Reinforcement Learning) can use numerical feedback to optimize heterogenous parameters (e.g. codes, hyper-parameters as well as tensors), though they are inefficient. In the experiments, we saw that \trace was more efficient than black-box methods by using a generalization of back-propagation. Finally, \trace can use rich feedback (e.g. language) to extract more signal for learning.

\paragraph{What can be traced?} 
\trace cannot convert all computational workflows into \opto problems. Workflows with recursive bundle operators or those requiring distributed/parallel computing are not compatible with the current implementation. A future work would be to expand the \trace implementation to support these scenarios.
In addition, the current implementation of \trace does not trace the execution within an operator defined by \bundle, though in principle this is possible.
There are also ambiguities in how an existing workflow can be traced and represented as a DAG. One example is when there is some sub-workflow following an \texttt{if} condition and another one following \texttt{else}. One can choose to wrap the entire code, including \texttt{if} and \texttt{else}, by \bundle as a single operator. On the other hand, one can also just wrap the sub-workflows and not trace the \texttt{if} condition nor represent it as part of the DAG. (That is, suppose the \texttt{if} condition is true; from the DAG, one cannot see the alternate path under \texttt{else}). The latter choice has a flavor of applying ``stop-gradient'' on the boolean condition, whereas the first choice enables back-propagation through also the logical condition. We summarized some design considerations in \cref{sec:constructing opto problem}. We foresee the choice of what to trace and how to trace in building Deep Agent Workflows will be an on-going research problem, similar to neural network architecture design.

\paragraph{Where do we get rich feedback?} 
The \opto framework captures an abstraction of rich feedback, called trace feedback (the execution trace and the output feedback), but it requires specification of operator descriptions and output feedback source to guide the optimization effectively. 
\opto can be more efficiently solvable than black-box problems only when the trace feedback provides information beyond reward signals; otherwise, information-wise, \opto is no easier than black-box problems.
The \trace framework automates the generation of the execution trace in \opto, when users of \trace decorate the workflow end-to-end. Currently, \algo uses the docstring of operators in the execution trace to understand the operators; nonetheless, \trace logs also the source code of the decorated methods, which can also be used in the future to design optimizers that uses more details of the operators.
In our experiments, we focus on output feedback that is automatically generated. In other contexts, e.g. users interacting with chatbots, we can natively gather natural language feedback or synthesize feedback~\citep{nie2023importance} based on raw observations.
We also anticipate that feedback in the form of images (e.g. users' gestures) or videos (e.g. videogame player showing a desired correction to agent's behavior) will be readily available and can be used as the output feedback in the \opto framework. 
While \trace can handle this feedback, the current design of \algo does not handle non-textual content (either in parameters, inputs or feedbacks). 
We anticipate future work along the lines of~\citep{nie2023importance} on enhancing feedback design and developing guidelines for designing more informative and directive optimizer and feedback mechanisms.

\paragraph{How to design adaptive optimizer for general \opto problems?}
The proposed \algo optimizer shows the possibility of designing a single optimization algorithm to solve a range \opto problems from diverse domains. However, we remark that \algo is just the first step; it is akin to the vanilla gradient descent algorithm, which shows a proof of concept but is not scalable for large problems.
The current design of \algo has several scalability limitations due to its summary approach that updates parameters through one call to an LLM. While we show in \cref{sec:numerical exp}, that \algo can compete with ADAM in small problems, \algo is not as computational efficient and cannot scale as well as ADAM; therefore, for large-scale numerical problems, \trace and \algo with the current design does not replace classical AutoDiff.
\algo is also limited by the ability of LLMs.
It has difficulty in handling parameters or nodes that cannot be compactly represented in text, which prevents it from optimizing neural network weights, or reasoning with large, stateful objects like a database.
Similarly, it likely cannot handle large graphs (with thousands of nodes) at the moment, as such a large graph would result in a huge context which may be beyond what LLMs can understand and reason about reliably.
This limitation is aggravated when dealing with noisy feedback or systems, as we need to present multiple graphs and feedback in the context at once. %
We also do not know how to rigorously define the concept of step size, which however we expect is important to handle noisy or local feedback.
We need further research on graph simplification and  representation, to reduce complexity and improve the efficiency of feedback propagation.
Lastly, right now \algo represents the constraints on parameters as part of the text description, but we have observed that LLMs do not always follow it. An effective workaround is to implement constraint checking in the workflow to throw exceptions (which are then handled as feedback). More specialized constraint handling techniques is an interesting research direction (e.g. ``projecting'' OPTO solver proposals for parameters onto their feasible sets), but they are not implemented in \algo yet.

\vspace{-1mm}
\section{Related Work}
\label{sec:related}
\vspace{-1mm}

\paragraph{Framework for Computational Workflows}
Frameworks such as LangChain~\citep{langchain-blog}, Semantic Kernels, AutoGen~\citep{wu2023autogen}, DSPy~\citep{khattab2023dspy} allow composing computational workflows and provide hand-engineered optimizers to tune an LLM's context (i.e. prompt templates, few shot examples, or tool libraries) using scalar feedback with black-box search techniques. 
They support tracing of the workflow to aid in profiling, debugging and visualization. 
In contrast, \trace uses tracing for \emph{automatic optimization}, and constructs a different representation of the computational graph which is suited for that purpose. Moreover, \trace is designed to be general-purpose and agnostic to the underlying frameworks of computational workflows users choose. In principle, one can apply \trace to decorate and tune a workflow based on a mix of Autogen, LangChain, DSPy codes. In fact, our experiments in~\cref{sec:expt_details} use workflows declared using both AutoGen and DSPy.

\paragraph{Optimization of Graphs of LLM Workflows}

There are multiple efforts to optimize the computational graph of LLM workflows, which is a special case of the \opto problem. These algorithms focus on optimizing prompts.
SAMMO~\citep{schnabel2024prompts} is an example for prompts that uses additional graph structure to make the optimizer efficient. SAMMO represents the prompt parameter itself as a program so as to enable more efficient black-box search through the space of programs.
DSPy \citep{ khattab2023dspy} can optimize directly the prompts or the few-shot examples to include using scalar reward feedback.
Retroformer \citep{yao2024retroformer} uses another small language model (LM) to provide suggestions/feedback (i.e. changing prompts) to improve the behavior of an actor LLM, where the small LM is tuned by offline RL.
Deep Language Networks~\citep{sordoni2024joint} view all of the prompts in an LLM workflow as tunable parameters and jointly optimizes them. They discovered that optimizing each parameter in isolation instead produces subpar results. %
\citep{zhuge2024language} frames LLM systems as graph where nodes are operations and edges are messages/connections. (Note that this is different from the DAG used in \trace; here nodes are messages and edges are input-output of operators) and optimizes for the connection on edges (binary variables) by REINFORCE using scalar reward feedback and prompts by LLMs . They optimize each component separately without considering each other; for example, the prompts are optimized individually without considering the graph topology or how they are used down the road. We suspect this approach can be less stable. Their prompt optimization part also does not take output feedback, but simply use an LLM to self-check whether the prompt meets the need of generating desired functions the user specified.
In contrast to these works, through the \opto framing, \trace supports \emph{joint} optimization of all parameters (prompts, hyperparameters, codes) with \emph{rich} feedback, and is agnostic to graph structures (e.g., changing these parameters can dynamically change the graph structure and connections between nodes). Users of \trace are free to specify which parameters they want to automatically optimize via online interactions.

\paragraph{LLM-Optimizers for Prompts and Codes}

There is a huge and fast growing literature on using LLMs as optimizers to improve prompts~\citep{pryzant2023automatic,zhou2022large,ma2024large,wang2023promptagent,tang2024unleashing} or codes~\citep{chen2023teaching,zelikman2023self,ma2024llamoco,ma2024large}. Different from the works mentioned above, here the focus has been on an isolated problem (e.g., changing the behavior of a single LLM or improving the code generation in the question-answering format) rather than considering a non-trivial workflow or agent with multiple components like above.
They do not consider optimizing prompts or codes as one component of a bigger workflow (e.g. implementing an autonomous agent), which is harder and requires the right  credit assignment.
In addition, these LLM-based optimizers, including OPRO~\citep{yang2024large}, often propose only principles of how prompts should be designed and requires crafting problem specific prompts (as opposed to a single optimization prompt that can be applied to different problems). For adapting them to new problems, users need to design new prompts.
\trace can also be applied to optimize trivial \opto problems where the returned graph has just a single node of the parameter (which are the scenarios considered by these works). Nonetheless, the main focus of this paper is to study how optimization can be done efficiently as the graph becomes nontrivial and for diverse applications.
\trace achieves this by using the abstract \opto problem framing. Since \opto encapsulates domain specific info in the graph, it enables designing fully instantiated optimizers that can be applied to multiple problems, rather than just principles which then requires hand crafting prompts for individual problems like previous works.

\paragraph{LLM-Optimizers for Hyperparameters}
Recent works like \citep{liu2024large,zhang2023using} use LLMs to optimize numerical hyperparameters, as an alternate to Bayesian optimization. Here in the experiments we show that \trace + \algo also can effectively learn hyperparamters, faster than Bayesian optimization.
The main difference between \trace and the aforementioned work is the representation of the problem. In \trace, we provide the graph to the LLM-based optimization (through the pseudo-algorithm representation), and we consume rich language feedbacks on the output, both of which accelerates hyper-parameter optimization.

\paragraph{\opto Related Setups}
\opto is a generalization of partial monitoring games~\citep{bartok2014partial}. 
If there exists a latent loss function that the feedback $f$ adheres to (e.g. as in~\citep{cheng2023llf}), those \opto instances can be written as partial monitoring game. However \opto admits a more general notion of feedback $f$, and we discuss solution concepts for them in~\cref{sec:opto solution concept}. %
On the other hand, \opto can be also viewed as a special case of Learning from Language Feedback (LLF) setup defined in \citep{cheng2023llf} with observations as the trace feedback. This is a framing of a \emph{meta} LLF problem. In the LLFBench Meta-World experiments of this paper (\cref{sec:expts}), we show \trace can be used to learn policy for LLF problems grounded to an application too.

\paragraph{AutoDiff and Back-propagation}
Back-propagation has been shown to be a very effective tool in optimizing differential computational workflows.
Our design of \trace is inspired by back-propagation and the ease of use of the AutoDiff framework PyTorch~\citep{paszke2019pytorch}. Nonetheless, we highlight that back-propagation (Backward Mode Differentiation) is not the only AutoDiff algorithm. For example, the gradient can be computed in a forward mode (Forward Mode Differentiation) as well, and there are also techniques of Checkpointing~\citep{hascoet2006enabling} and Truncated Back-Propagation approximation~\citep{shaban2019truncated} for efficiency. What are the equivalent ideas of these methods for general computational workflows? We think this is an interesting future research direction.

\section{Examples of \opto} \label{sec:more opto examples}

To ground the \opto setup, we show how \opto is related to some existing problems with examples.
\begin{example}[Neural network with back-propagation]
The parameters are the weights. $g$ is the neural computational graph and $f$ is the loss. An example context $\omega$ can be ``Minimize loss''.
The back-propagation algorithm, in view of the \opto formulation, is embedded in the \opto optimizer. For example, an \opto optimizer here is a composition of back-propagation and gradient descent, where back-propagation takes $\tau$ to compute the propagated gradient at the parameter. %
\end{example}
\begin{example}[Code Debugging]
The parameters are the codes.  $g$ denotes the stacked trace and $f$ is the error message returned by a compiler. $\omega$ can be ``Make no error''.
\end{example}
\begin{example}[RL]
The parameters are the policy. $g$ is the trajectory (of states, actions, rewards) resulting from running the policy in a Markov decision process; that is, $g$ documents the graphical model of how an action generated by the policy, applied to the transition dynamics which then returns the observation and reward, etc. $f$ can be the termination signal or a success flag.  $\omega$ can be ``Maximize return'' or ``Maximize success''.
\end{example}
\begin{example}[Hyperparameter Tuning of ML Pipeline]
The parameters are e.g. learning rates and architectures. $g$ describes the stages of the ML pipeline and the evaluation on the validation set, and $f$ is the validation loss. $\omega$ can be ``Minimize validation error''.
\end{example}
\begin{example}[Prompt Optimization of an LLM Agent]
The parameters are the prompt of an LLM workflow. $g$ is the computational graph of the agent and $f$ is the feedback about the agent's behavior (which can be scores or natural language). $\omega$ can be ``Maximize score'' or ``Follow the feedback''.
\end{example}
\begin{example}[Multi-Agent Collaboration]
The parameters are each agent's prompts. $g$ describes the entire conversation flow between agents, and $f$ is the feedback about whether the task is successful after each agent performs their action. $\omega$ can be ``A group of agents coordinate to finish a task.''.
\end{example}

As mentioned, the computational graph $g$ returned by the Trace Oracle $\TT$ may have different graph structures. The length of the execution trace, e.g., in the debugging example above depends on how far the code executes. Similarly, the rollout length of in the RL problem can be randomly determined. The formulation of the Trace Oracle abstracts the details of a computational workflow, so problems from different domains can be framed in the unified framework. This abstraction allows us to design the computational tool \trace for various applications.

\section{\trace Handles Error in Execution as Feedback}  \label{sec:trace error}
It is worth mentioning that execution error can be directly used as feedback to optimize parameters in \trace.
When execution error happens within a method decorated by \bundle, \trace would adds a special exception node to the global computational graph and throw an \exception to stop the computation.
The computational graph ends at where the execution error happens. This exception node becomes the new output of the inputs to the decorated method (since the original method raises an error) and is the output of the truncated computational graph. Messages in \exception can then be used as the feedback $f$ in \opto and propagated from the exception node to the parameters. By calling an \opto optimizer, the parameters can be updated to avoid causing the same execution error. See the exception handling code in \cref{fig:overall_pipeline}.

We find that this error handling mechanism has two convenient usages. First, this allows using \trace to automatically debug issues in the workflow due to incorrect parameter settings. Such errors can happen frequently especially when codes are parameters, as during optimization codes not satisfying syntax or downstream API requirements can happen. The second usage is to enforce constraints the workflow has to satisfy at different stags of computation. With \trace, if an intermediate computed result does not satisfy the constraint, we can simply throw an exception which states the desired constraint. This error signal would be caught by \trace and can then provide early feedback to efficiently improve the parameters, since the graph is truncated at the error.

\section{Analysis of \trace} \label{sec:trace analysis}

\subsection{Proof of Complexity}

\cref{alg:subgraph propagator} propagates the subgraph, represented by a priority queue (implemented as a min-heap). At a time, it needs to maintain the subgraphs coming from $W$ children separately. This leads to the space complexity of $O(WN)$.
This $O(WN)$ space complexity leads to the extra $W N\log N$ factor in the time complexity of \graphpropagator compared with back-propagation, which is the time needed for merging $W$ subgraphs of size $O(N)$.

\subsection{Proof of Lower bounds} \label{sec:proof of lower bound}

Consider an \opto problem whose goal is to find a parameter matching a $k$-digit binary number. The computation checks each digit against a reference number in an arbitrary order. The feedback is either ``$N^{th}$ check failed'' or ``All checks succeeded''. Propagated feedback must communicate $k$ bits of information to interpret the feedback correctly; and the minimal subgraph conveys exactly that information. Updating the parameter using the minimal subgraph is trivial, whereas without it there are $2^k$ possibilities to check.

\section{When is \opto Efficiently Solvable?}  \label{sec:opto solution concept}

We show that \opto covers a wide range of complicated optimization problems. This shows that if \opto can be efficiently solved, then many complex workflows can be efficiently optimized. However, the generality of \opto also raises some fundamental questions, such as if \opto is well defined and when \opto can be efficiently solved.
These questions stem from its generality of the context $\omega$ and the output feedback $f$ in \opto, since e.g. they can be anything descriable texts. This flexibility makes the scope of \opto go beyond standard mathematical optimization problems, where a setup has a fixed context $\omega$ (e.g., ``First-order optimization'') and a fixed type of output feedback $f$ (a descent direction). Fully characterize the properties of \opto, due to its generality, is beyond the scope of this paper and would require years of future research to come. Nonetheless, here we attempt to provide some preliminary answers and point out some research questions.

\subsection{What is a solution?}
Classical mathematical optimization problems have a problem definition which itself is the solution concept. For example, in a minimization problem, it is clear we want to find the minimum of an objective function; even for problems as abstract and general as an equilibrium problem, the problem setup clearly states the solution concept of finding a point/set satisfying an equilibrium inequality~\citep{cheng2020online}. One common pattern of these problems is that the solution concept is something that can be described as conditions on feedback that the parameter should satisfy.

By contrast, in a \opto problem $(\Theta, \omega, \TT)$,  by varying the context $\omega$, the desired parameter can
change from one extreme to another. For example $\omega$ may state ``Follow the feedback'' or ``The feedback is adversarial.''. Therefore, we need define the solution concept of \opto differently, rather than just using the feedback. We need to also consider the context $\omega$ appropriately.
Below we make an attempt to give an axiom of \opto for its solution to be well defined.
\begin{axiom}[Verifiability] \label{as:verifiability}
There is an verification oracle (a human, a machine learning model, or a polynomial-time algorithm) when given $(\theta, \omega, f)$ can verify whether $\theta$ is a solution or not.
\end{axiom}
Notice the verification oracle in \cref{as:verifiability} is not limited to just algorithms. This is intentional because we currently do not have algorithms that are intelligent enough to process the wide range of contexts and feedback that \opto allows. Therefore, we include human judgement or the use of LLMs or other AI systems as part of the definition, while acknowledging the impreciseness of the statement due to \opto's soft computing nature. Lastly we note the verifiability is only defined with respect to the context $\omega$ and the output feedback $f$, not the execution trace $g$. That is, the verification of a solution depends only on the output of computation.

\subsection{Does a solution exist?}
Under \cref{as:verifiability}, we can start to ask the basic question of whether a solution to an \opto problem exists or not. There are clearly problems where no solution exists (that is, no parameter in $\Theta$ can be verified by the verification oracle).
For example if the feedback $f$ is contradicting and yet the context $\omega$ is ``Follow the feedback.'', then there would be no solution that is satisfactory. On the other hand, if $\omega$ is ``Ignore the feedback'', all parameters can be solutions.
In the following, we assume solutions of \opto under consider exist. This assumption would rule out problems, e.g., where the feedback is adversary to the context, and  makes solving \opto is a well-defined search problem.
\begin{assumption}
For an \opto problem $(\theta,\omega,\TT)$, we assume there is at least a parameter $\theta\in\Theta$ such that it can be verified as a solution by the verification oracle.
\end{assumption}

\subsection{Can \opto be efficiently solved?}
So far our discussion establishes \opto as a well-defined search problem, based on qualification on the context $\omega$ and the output feedback $f$. %
However search problems can be NP-hard. In other words, we know that,  without the execution trace, there are search problem instances modeled by some $\omega$ and oracle giving $f$ that cannot be efficiently solved.
Take RL for a tabular MDP as an example of \opto problem. Without the execution trace (i.e., not seeing the Markovian structure and trajectories), the problem has an exponential complexity (due to the size of the policy space) and we know by using the execution trace here, tabular RL can be solved approximately in polynomial time~\citep{kakade2003sample}. %
Another example is training of neural networks. Without the execution trace, we have a complex black-box optimization with a loss value, without gradients, whereas an execution trace allows implementation of back-propagation to compute the gradients at the parameters.

More broadly speaking, if we consider a ``human'' as an optimizer for \opto, we see that (expert) engineers/researchers, when equpped with additional computational tools, can efficiently solve a broad range of \opto problems (such as by using the execution trace.
From these observations, we conjecture using information in the execution trace is the key to unlock efficient \opto.
More precisely, we conjecture that \opto is efficiently solvable when the context and the trace feedback need to provide information to construct a corrective search direction.
For example, when the output feedback back is just a scalar loss, and yet the context + execution trace feedback does not provide enough information to compute a descending direction then \opto reduces back to a black box problem. (See the problem instance in \cref{sec:proof of lower bound}).
Nonetheless, identifying which subsets of \opto are efficiently solvable is a big open research question.

\section{Additional Details of Trace and \algo}
\label{sec:algo_detail}

\subsection{Backward Step of \trace}

The \graphpropagator extracts the minimal subgrpah of the full computational graph of the workflow. Here we show a visualization using the example in \cref{fig:code repr}.

\begin{figure}[ht]
\centering
\begin{subfigure}[b]{0.49\textwidth}
\includegraphics[width=0.9\textwidth]{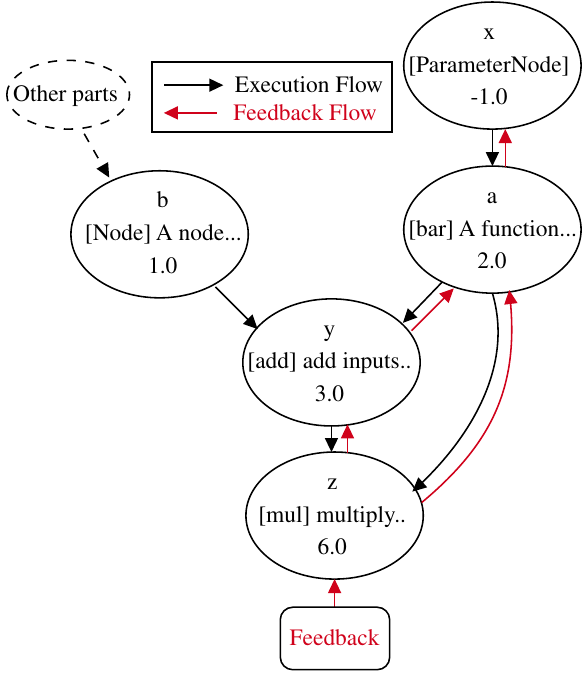}
\caption{This is an illustrative example of the graph constructed by \trace and how feedback is backpropagated to the parameter \texttt{x}.}
\end{subfigure}
~
\begin{subfigure}[b]{0.45\textwidth}
\begin{lstlisting}[language=Bash,breaklines=true,showstringspaces=false,basicstyle=\fontsize{7}{7}\ttfamily, numbers=none]
#Code:
a = bar(x)
y = add(b, a)
z = mul(a, y)

#Definitions:
[mul] This is a multiply operator
[add] This is an add operator.
[bar] This is a method that does negative scaling.

#Inputs:
b=1.0

#Others:
a=2.0
y=3.0

#Output
z=6.0

#Variable
x=-1.0

#Feedback:
Output should be larger.
\end{lstlisting}
\caption{We create a succinct summary of the computation graph using a language that mimics a program.}
\label{fig:opt-summary appendix}
\end{subfigure}
\caption{\textbf{Optimization Representation}. For a program of \texttt{x = node(-1.0); a = bar(x); y = a + 1; z = a * y} and the optimization objective of $\max_x z$, \trace automatically constructs a computation graph and represent the optimization problem as a debugging report. Note that the real program and the traced execution graph are different.}
\label{fig:optimization}
\end{figure}

\subsection{Prompts used in \algo}
\label{sec:optoprime_prompts}

\algo is an LLM-based optimizer. Its prompt is composed of the following parts.
\begin{enumerate}
    \item System Prompt: Representation Prompt (\cref{fig:representation prompt}) + ReAct+CoT Output Prompt (\cref{fig:format prompt})
    \item User Prompt (\cref{fig:user prompt without memory} or \cref{fig:user prompt with mem})
\end{enumerate}
where + denotes concatenation.
We list the prompt templates of different components below.

\begin{figure}[ht]
    \centering
    \begin{lstlisting}[escapechar=!]
    You're tasked to solve a coding/algorithm problem. You will see the !instruction!, the !code!, the !documentation! of each function used in the !code!, and the !feedback! about the execution result.

    Specifically, a problem will be composed of the following parts:
    - #Instruction: the !instruction! which describes the things you need to do or the question you should answer.
    - #Code: the !code! defined in the problem.
    - #Documentation: the !documentation! of each function used in #Code. The explanation might be incomplete and just contain high-level description. You can use the values in #Others to help infer how those functions work.
    - #Variables: the input !variables! that you can change.
    - #Constraints: the !constraints! or descriptions of the !variables! in #Variables.
    - #Inputs: the values of other !inputs! to the !code!, which are not changeable.
    - #Others: the intermediate values created through the !code! execution.
    - #Outputs: the result of the !code! output.
    - #Feedback: the !feedback! about the !code!'s execution result.

    In #Variables, #Inputs, #Outputs, and #Others, the format is:

    <data_type> <variable_name> = <value>

    If <type> is !(code)!, it means <value> is the source !code! of a python !code!, which may include docstring and definitions.
    \end{lstlisting}
    \caption{Representation Prompt that phrases the \opto update as a pseudo-algorithm question.}
    \label{fig:representation prompt}
\end{figure}

\begin{figure}[ht]
    \centering
    \begin{lstlisting}[escapechar=!]
    Output_format: Your output should be in the following json format, satisfying the json syntax:

    {{
    "reasoning": <Your reasoning>,
    "answer": <Your answer>,
    "suggestion": {{
        <variable_1>: <suggested_value_1>,
        <variable_2>: <suggested_value_2>,
        }}
    }}

    In "reasoning", explain the problem: 1. what the #Instruction means 2. what the #Feedback on #Output means to #Variables considering how #Variables are used in #Code and other values in #Documentation, #Inputs, #Others. 3. Reasoning about the suggested changes in #Variables (if needed) and the expected result.

    If #Instruction asks for an answer, write it down in "answer".

    If you need to suggest a change in the values of #Variables, write down the suggested values in "suggestion". Remember you can change only the values in #Variables, not !others!. When <type> of a variable is !(code)!, you should write the new definition in the format of python !code! without syntax errors, and you should not change the function name or the function signature.

    If no changes or answer are needed, just output TERMINATE.
    \end{lstlisting}
      \caption{ReAct+CoT Output Prompt that instructs LLMs should respond in the format of (reasoning, answer, suggestion) and explains the output format.}
    \label{fig:format prompt}
\end{figure}

\begin{figure}[ht]
    \centering

    \begin{lstlisting}
    Now you see problem instance:

    ================================
    {actual_problem_instance}
    ================================

    Your response:
    \end{lstlisting}
    \caption{User Prompt for \algo without Memory}
    \label{fig:user prompt without memory}
\end{figure}

\begin{figure}[ht]
    \centering
    \begin{lstlisting}[escapechar=!]
    Now you see problem instance:

    ================================
    {actual_problem_instance}
    ================================

    Below are some !variables! and their feedbacks you received in the past.

    {
        "!variables!": {
            {variable1_name}: {variable1_value1}
            {variable2_name}: {variable2_value1}
            ...
        },
        "!feedback!": {feedback_1}
    }

    {
        "!variables!": {
            {variable1_name}: {variable1_value2}
            {variable2_name}: {variable2_value2}
            ...
        },
        "!feedback!": {feedback_2}
    }
    ...


    Your response:
    \end{lstlisting}
    \caption{User Prompt for \algo with Memory}
    \label{fig:user prompt with mem}
\end{figure}

\begin{figure}[ht]
\centering

\begin{lstlisting}
#Instruction
{instruction}

#Code
{code}

#Documentation
{documentation}

#Variables
{variables}

#Constraints
{constraints}

#Inputs
{inputs}

#Others
{others}

#Outputs
{outputs}

#Feedback:
{feedback}
\end{lstlisting}
\caption{Problem Template used to fill the User Prompt. By default the Instruction (which is the context $\omega$ of \opto) is ``You need to change the <value> of the variables in \#Variables to improve the output in accordance to \#Feedback.''
}
\label{fig:problem template.}
\end{figure}

\clearpage
\

\section{Comparison between \trace and TextGrad}
\label{sec:textgrad_comparison}

Beyond the empirical results described in~\cref{sec:exp of textgrad}, there are several first principles reasons to prefer the \trace design to build future LLM-powered generative optimizers.

\paragraph{Joint vs. Individual Parameter Optimization:}
When there are multiple parameters to be optimized in a workflow, \trace takes a ``joint optimization'' view (constructing the minimal subgraph involving all of the parameters described in the initialization of \algo). This flexible design also allows an alternative approach of ``individual optimization'' akin to co-ordinate descent if desired, i.e., fix all but one parameter and individually optimize each parameter. Such an optimization heuristic can be accomplished, e.g., by instantiating an \algo instance per parameter and iterating through their updates; however, for many problems (like the toy example below), this strategy, which TextGrad employs, can be sub-optimal because a consistent optimizer must additionally maintain state about any suggested updates to all other parameters when reasoning about an individual parameter update.

\begin{lstlisting}[language=Python, caption=Individual Parameter Optimization is suboptimal in this example compared to \trace.]
@bundle()
def function1(x):
    return x > 0
@bundle()
def function2(y):
    return y %
def xor_test(x, y):
    return function1(x).neq(function2(y))
input1 = node(3, trainable=True); input2 = node(4, trainable=True)
xor_test(input1, input2).backward(feedback="Find a set of inputs to make the return True.")
\end{lstlisting}

\paragraph{Heterogenous Parameters:} \trace encapsulates all Python datatypes as \verb|node|, including numbers, floats, strings, etc. In contrast, many generative optimization libraries like TextGrad restrict their ``variables'' (i.e. parameters and intermediate values) to only be strings; consequently running numerical optimization experiments, or optimizing code alongwith prompts and hyper-parameters can be challenging to set up.

\paragraph{Modularity:} \trace provides a clean separation between the infrastructure (a platform akin to PyTorch) and a generative optimizer (\algo akin to the Adam optimizer). Trace provides an object ``TraceGraph'' that describes the computational graph. This graph is independent of optimizers and LLMs. Such independent computation graph objects do not exist in TextGrad. We hope the \trace-provided interface can spur the development of many other generative optimizers, like how the OpenAI Gym API allowed the development of several Deep RL algorithms targeting a unified representation of RL problems.

\paragraph{Flexible Representation:} The \verb|bundle| functionality of \trace is more flexible in representing computations to an LLM than pre-defined templates used in other optimization frameworks. TextGrad relies on special functions to chain variables together, such as using ``FormattedLLMCall'', while Trace supports any user-defined functions through decorators like \verb|bundle|.

\paragraph{Scalability:}
The strategy taken by \algo can scale better to larger graphs (both in terms of error accumulation across multiple LLM calls, and the costs of multiple LLM calls) compared to requiring one LLM call per computation node (and we see with \verb|bundle| that not all computation nodes present equal difficulty for an optimizer). The difference in terms of computation costs between the two strategies can be large, especially when not all operators in the graph are as complex as querying LLM.
However there may be even better optimizers that blend \algo and TextGrad so as to divide-and-conquer huge computation graphs, we are excited about this avenue for future research.

\section{Experiment Details}
\label{sec:expt_details}

\subsection{Token Counts} \label{sec:token counts}

We include token counts for the OPRO and OptoPrime prompts used across all our experiments, at the first iteration of optimization (note that OPRO's token usages grows \emph{linearly} with iterations).
We can see that indeed OptoPrime consumes significantly more tokens than OPRO. However, we observe consistently that even allowing 7-10x more iterations of OPRO so as to equalize token costs, the OPRO performance plateaus to a worse level than OPTOPrime (e.g. Figure 1: OPRO at Iter 7 vs. OptoPrime at Iter 2; Figure 5b: OPRO at Iter 50 vs. OptoPrime at Iter 5; Figure 6b: OPRO at Iter 30 vs. OptoPrime at Iter 10, etc.). OPRO is suboptimal not due to a token limit but instead a lack of information, which is captured and represented using Trace.

\begin{table}[h]
    \centering
\begin{tabular}{@{}c|c|c@{}}
\toprule
     Domain & OPRO & \algo  \\
     \midrule
    Numerical Opt &  175 & 918 \\
    BigBench-Hard &  400 & 1883 \\
    Traffic Opt & 198 & 1679 \\
    MetaWorld & 470 & 7101 \\
    Battleship & 437 & 1305 \\ \bottomrule
\end{tabular}
\vspace{1mm}
\caption{Token counts.}\label{tab:token_counts}
\vspace{-5mm}
\end{table}

\subsection{Battleship}
\label{app:sec:battleship}

We implement a simple battleship game board in Python. The exact code is in the supplement. The game offers a string-based visualization of the board. It randomly places different types of ships on a 2-dimensional board with pre-specified width and height when it initializes. The agent does not see the ship location and has to select a coordinate on the board to hit next. One additional rule of this game is that the agent can go again if their previous coordinate selection (fire) is a hit, not counting as the finish of a turn. In Figure~\ref{fig:battleship-learning}, we ran 10 trials, where in each trial, we ran 20 iterations of training. We measure the reward as \% of ship squares hit (over all squares occupied by ships). The reward plateaued at 60\% because the game has a chance element (heuristics and strategies can only go so far -- strategy is only in effect if a hit happens. Otherwise, there is no information about where ships might be).

\subsection{Numerical Optimization}
\label{app:sec:numer-opt}

Any classical numerical optimization problem can be framed as an \opto problem. Consider $h(x)$ and a target $y^*$, in a context $\omega$ finding the $y^*$ by changing $x$; we know the most useful corrective $f$ feedback to change $x$ is the gradient $\nabla_h x$. Similar to \trace, AutoDiff packages like PyTorch's AutoGrad have implemented dynamic graph construction with special classes like \texttt{torch.Tensor}. We want to validate whether it is possible to rely on binary text feedback, a graph automatically constructed by \trace, and \algo to update $x$ in the context of minimizing $|y - y^*|$.

We constructed a synthetic task environment where we can create a complex computation graph with arbitrarily complex connections between numerical variables. The focus of this environment is on the complexity of the graph, not on the complexity of the numerical operators. Therefore, we only use one-dimensional input and basic arithmetic operators to create a numerical optimization problem solvable by a first-order optimizer. This environment constructs a computational graph by sampling a number of times. At each time, it will either use a previously computed variable or sample a new variable, and an operation will be sampled to combine them. The optimization task is, for a fixed number of steps, an optimizer needs to output $x$ that minimizes $y$.

We evaluate the following baseline methods. Basic Agent: a basic LLM agent that simply stores past information of $(x_{t-1}, y_{t-1})$ in context before choosing the next $x_t$. OPRO Agent: a basic LLM agent but we implement the state-of-the-art LLM optimizer OPRO~\citep{yang2024large}, which updates the meta-prompt of the basic LLM agent. Torch + Adam: the problem we construct is end-to-end differentiable. Therefore, we simply pass in \texttt{torch.Tensor(x)} as input and use Adam optimizer to update. We tune the learning rate slightly and found 1e-1 to work well.
We compare two kinds of Trace-based optimizers: Trace, where we allow \algo to read in the entire computation graph before updating $x$, or Trace Masked, where we hide the computation graph.

We run 30 trials over different computation graphs and start all methods with the same initial $x, y^*$. We compute the absolute error, which is $|y - y^*|$. On average, Trace is able to match the best-in-class first-order gradient optimizer Adam~\citep{kingma2014adam}. It is not entirely surprising that all the other baselines are performing worse due to a lack of access to the computation graph. To our surprise, OPRO, by only accessing the history of input and output, as well as changing the meta-prompt, is able to eventually discover the correct solution. This confirms why there were early signs of success using LLMs for black-box optimization in a simple plug-and-play style. However, OPRO is not an efficient optimizer because it lacks access to the \trace oracle. We show OPRO struggles even more when the computation graph gets more complex.

\subsection{Traffic Control}
\label{app:sec:traffic}

We tested \algo in a traffic control problem which is an instance of hyper-parameter tuning. We used UXSim~\citep{seo2023uxsim} to simulate traffic at a four-way intersection, where the tunable parameters are the duration of the green lights for each direction of traffic flow.
The feedback is a scalar loss calculated by monitoring the flow of a pseudo-random sequence of vehicles arriving at the intersection over a period of $30$ minutes.
The loss computes an estimate of the delay experienced by all vehicles due to the intersection, as well as variability in this estimate for every link in the network; lower values are better. The goal of an optimizer is to identify values for all of the green light duration so as to minimize the loss using the fewest number of traffic simulations.
If the green light duration for a given traffic flow direction is set too low, then vehicles will queue up over time and experience delays, thereby lowering the score for the intersection. However, if the green light duration for a given direction is set too high, vehicles in other directions will queue up and experience delays, thereby lowering the score for the intersection. Hence an optimizer must find the right trade-off for temporally distributed and variable demands.

In Figure~\ref{fig:Traffic_expts} we report the performance of a SOTA heuristic from the traffic control literature, SCATS~\citep{scats} (adapted to this toy setting) as well as two black-box optimization techniques: Gaussian Process Minimization (GP)~\citep{frazier2018bayesian} and Particle Swarm Optimization (PSO)~\citep{kennedy1995particle}. All methods are initialized to evaluate the same starting parameter. GP and PSO further evaluate $5$ random parameters; moreover, if they query a previously evaluated point, that query is replaced by a randomly sampled parameter.
GP constructs a surrogate model to mimic the black-box traffic simulation function which maps from parameters to observed score. Then it minimizes a utility function (e.g. the lower confidence bound) using the surrogate model to pick the next parameter to evaluate.
PSO on the other hand maintains $5$ particles in parameter space, each with a position and velocity.
At each iteration of PSO, particles update their positions according to their previous positions and velocity, evaluate the function at the updated positions, and update the velocities of all particles using the observed values.
Although GP and PSO are both black-box methods, GP can be thought to replace \trace oracle with instead a smooth differentiable surrogate function; whereas PSO is very different and maintains a candidate set of parameters (can be thought of as conceptually related to \algo with memory).

GP appears to be bad because even when it converged, the exploration heuristic randomly samples parameters rather than pick the converged parameter. PSO appears bad because $10$ iterations is insufficient for its convergence. Note that given enough number of iterations, black-box approaches will eventually perform well.
\trace is quickly competitive with the SCATS heuristic, whereas OPRO is not. Moreover, we find that memory is crucial for \algo to perform well at this task. Finally, \trace consumes additional overhead compared to black-box methods; beyond the space and time complexity for running the traffic simulation, \trace additionally materializes the computation graph per iteration. Thus it can also be more expensive per LLM call compared to OPRO.

\subsection{BigBench-Hard}
\label{app:sec:bigbench}

\begin{figure*}
\centering
\begin{subfigure}[t]{0.5\textwidth}
\begin{lstlisting}[language=Python,breaklines=true,showstringspaces=false,basicstyle=\fontsize{6}{6}\ttfamily]
import trace

class Predict(trace.Module):
    def __init__(self):
        self.prompt_template = trace.node("""
        Given `question`, produce the `answer`.
        question: {}
        answer:
        """,
        trainable=True)

    def forward(self, question):
        user_prompt = self.create_prompt(question)
        response = self.call_llm(user_prompt)
        answer = self.extract_answer(question, response)
        return answer

    @trace.bundle(trainable=True)
    def create_prompt(self, question):
        """formulate the prompt with the question"""
        return self.prompt_template.format(question)

    @trace.bundle(trainable=True)
    def extract_answer(self, question, response):
        """Extract the answer out of LLM response"""
        answer = response.split("answer:")[1].strip()
        return answer
\end{lstlisting}
\caption{We write a workflow that prompts an LLM for a question and extracts the answer.}
\end{subfigure}
~
\begin{subfigure}[t]{0.45\textwidth}
\begin{lstlisting}[language=Python,breaklines=true,showstringspaces=false,basicstyle=\fontsize{6}{6}\ttfamily, escapechar=!]
import trace

dataset = load_dataset(link, task)
predict = Predict()
optimizer = trace.Optimizer(predict.parameters())

for ex in dataset:
    # Forward pass
    response = dp.forward(ex['question'])
    if response == ex['answer']:
        !feedback! = "The answer is correct"
    else:
        !feedback! = "The answer is incorrect"

    # Backward pass and update
    optimizer.zero_feedback()
    optimizer.backward(response, !feedback!)
    optimizer.step()

\end{lstlisting}
\caption{The optimizer class takes in any parameter regardless of whether it is code or text. Although the actual optimization implementation can provide different treatments to many input types, the user interface stays consistent.}
\end{subfigure}
\caption{\textbf{LLM-based Workflow Optimization Example.}}
\label{fig:prompt}
\end{figure*}

\begin{table}[ht]
\footnotesize
\centering
\begin{tabular}{@{}l|ll|ll|ll|@{}}
\toprule
Task Name                               & DSPy  & +CoT & DSPy-PO & +CoT & Trace & +CoT \\ \midrule
tracking shuffled objects $7$ objects & 37.39 & 90.0     & 90.43   & 90.43       & 37.8  & 87.8      \\
salient translation error detection     & 51.3  & 70.87    & 51.3    & 69.57       & 63.0  & 70.0      \\
tracking shuffled objects $3$ objects & 39.13 & 94.35    & 97.39   & 93.91       & 38.7  & 96.5      \\
geometric shapes                        & 50.43 & 62.17    & 59.13   & 60.43       & 49.6  & 62.2      \\
object counting                         & 0.0   & 40.0     & 0.0     & 74.35       & 42.2  & 80.4      \\
word sorting                            & 0.0   & 0.0      & 0.0     & 0.43        & 84.3  & 74.3      \\
logical deduction five objects          & 70.0  & 80.43    & 70.0    & 74.78       & 48.7  & 75.7      \\
hyperbaton                              & 74.78 & 86.52    & 74.78   & 88.26       & 78.3  & 91.7      \\
sports understanding                    & 0.0   & 0.0      & 0.0     & 0.0         & 79.6  & 45.7       \\
logical deduction seven objects         & 68.7  & 64.78    & 68.7    & 64.78       & 45.7  & 69.6      \\
multistep arithmetic two                & 0.0   & 93.04    & 0.0     & 93.04       & 94.8  & 88.7       \\
ruin names                              & 84.35 & 87.83    & 84.35   & 87.83       & 87.8  & 90.0      \\
causal judgement                        & 7.78  & 70.66    & 74.25   & 70.66       & 70.1  & 54.5      \\
logical deduction three objects         & 85.22 & 97.39    & 85.22   & 97.83       & 91.7  & 97.0      \\
formal fallacies                        & 1.74  & 81.3     & 62.17   & 81.3        & 73.5  & 67.8      \\
snarks                                  & 86.08 & 87.34    & 86.08   & 87.97       & 81.6  & 87.3      \\
boolean expressions                     & 0.0   & 98.26    & 64.35   & 98.26       & 88.7  & 96.5      \\
reasoning about colored objects         & 53.04 & 91.3     & 89.13   & 91.3        & 91.3  & 95.7      \\
dyck languages                          & 0.0   & 8.7      & 7.83    & 8.7         & 26.5  & 9.6       \\
navigate                                & 0.0   & 95.65    & 0.0     & 97.39       & 59.6  & 92.6      \\
disambiguation qa                       & 67.83 & 66.09    & 73.91   & 66.09       & 75.7  & 59.1      \\
temporal sequences                      & 99.57 & 99.13    & 97.39   & 99.13       & 97.8  & 98.3      \\
web of lies                             & 0.0   & 0.0      & 0.0     & 0.0         & 49.6  & 90.4      \\
tracking shuffled $5$ objects  & 37.83 & 96.09    & 37.83   & 96.09       & 58.3  & 88.7      \\
penguins in a table                     & 69.84 & 92.86    & 97.62   & 92.86       & 81.7  & 91.3      \\
movie recommendation                    & 83.48 & 76.09    & 83.48   & 76.09       & 81.3  & 75.7      \\
date understanding                      & 69.13 & 85.65    & 69.13   & 85.65       & 70.4  & 85.7      \\ \bottomrule
\end{tabular}
\vspace{2mm}
\caption{Big Bench-Hard Per-Task Result. 0-shot performance. Some 0.0 here shown is because DSPy cannot find the clean/stripped output that matches what the automatic evaluation method expects. With additional human engineering, these numbers can improve.}
\end{table}

Perhaps more surprisingly, there are many components that a workflow needs to learn. Some of these components can be the prompt to generate output from an LLM, while other components can be code that needs to further process these outputs. In many workflows today, enabled by LangChain~\citep{langchain-blog} and DSPy~\citep{khattab2023dspy}, only a small part of this workflow, predominantly, the input to an LLM API call, is optimized. These libraries optimize input to an LLM, and human engineers process that input and integrate it into other systems. Indeed, both libraries can enable robust and swift large-scale engineering efforts to build LLM-based software. However, if our goal is to develop self-adapting agents that can modify their own behavior, we should not ignore one of LLM's greatest strengths: code generation. \trace allows us to unify prompt optimization and code generation, which enables the creation of agents capable of fast learning.

In this example of an LLM-based workflow (Figure~\ref{fig:prompt}), there are three parameters that are flagged as \verb|trainable| for the optimizer: \verb|prompt_template|, \verb|create_prompt|, and \verb|extract_answer|. Note that two of them require the LLM to generate Python code, and one of them requires the LLM to modify a text. \trace abstracted away the different data types and enabled direct update and optimization of them.
Furthermore, a human engineer is often tasked with writing an error-free \verb|extract_answer|. The output of an LLM can be highly stochastic and can often change over time; the code that is used to extract the response of an LLM has to be extremely robust and, therefore, arduous to create. Whenever a major distribution shift happens in the LLM output, this code needs to be rewritten by a human engineer, and it is hard for humans to anticipate all of LLM's output patterns.

We set up the task of end-to-end workflow optimization. Unlike a typical LLM benchmark evaluation, where a lot of effort went into creating the perfect \texttt{evaluate(answer, target)} method so that all kinds of LLM outputs were post-processed, cleaned, and formatted to match the ground truth, we choose a simple evaluation function (that extracts a segment or does exact string matching) and place the burden on the workflow itself to figure out how to create the right answer to satisfy the evaluation metric. We choose Big-Bench Hard~\citep{suzgun2022challenging} as our task because it has 23 subtasks and contains both language and algorithmic tasks.

We split each task dataset into training, validation, and test. For Trace and Trace-CoT, we use the first 15 examples for training, 5 examples for validation (picking the best learned workflow), and then evaluate the performance on test examples. We use template-based positive and negative feedback during training. The positive feedback is ``The answer is correct! No need to change anything.'' The negative feedback is ``The answer is wrong. We expect the output of your answer to be \{ANSWER\}. Please modify the prompt and relevant parts of the program to help LLM produce the right answer.'' DSPy's prompt optimization method does not explicitly require a validation set, therefore, we just used all 20 examples for training. For both, we only optimize for 1 epoch. We either start with the vanilla boilerplate prompt template used by DSPy or we use the slightly sophisticated template used by DSPy's CoT module. Trace optimizes both DSPy's original design and outperforms their own optimizer COPRO by 10\% on algorithmic tasks.

Big-Bench Hard requires different answer outputs. Out of 23 tasks, 14 tasks require a multiple-choice answer with options provided in the question. 4 of them require yes/no. 1 task requires True/False, while 1 task requires valid/invalid. And the 3 remaining tasks require answers that contain words or numbers. Even though DSPy's meta-prompt optimization is trained on each task individually, the output of LLM to the evaluation method is still not post-processed, resulting in low performances of these tasks. However, Trace can optimize code and LLM prompt jointly to successfully deliver the response expected by an automatic evaluation method.

\subsection{LLFBench Meta-World}
\label{app:sec:metaworld}

\begin{figure}
\centering
\begin{subfigure}[t]{0.11\textwidth}
\centering
\vspace{-22mm}
\includegraphics[trim={10.8cm 3cm 10.8cm 3cm},clip,width=\textwidth]
{figures/mw_figures.png}
\end{subfigure}
\begin{subfigure}[t]{0.87\textwidth}
\centering
\begin{subfigure}[t]{0.32\textwidth}
\includegraphics[width=\textwidth]{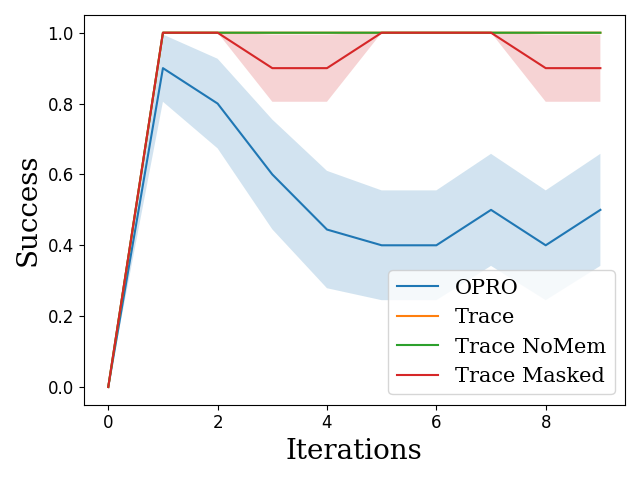}
\caption{Reach (Train)}
\end{subfigure}
\begin{subfigure}[t]{0.32\textwidth}
\includegraphics[width=\textwidth]{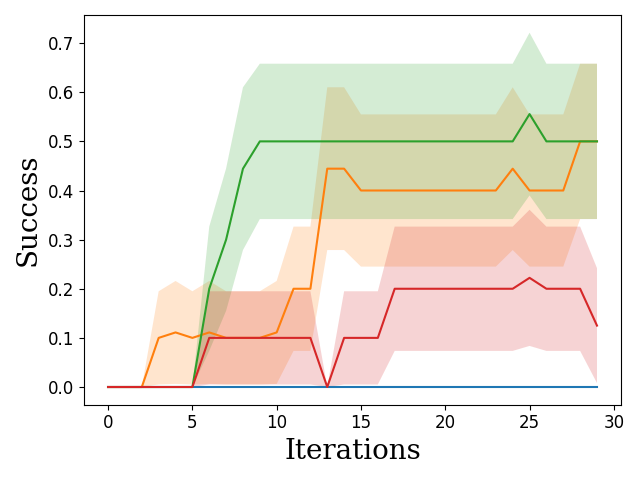}
\caption{Pick-place (Train).}
\end{subfigure}
\begin{subfigure}[t]{0.32\textwidth}
\includegraphics[width=\textwidth]{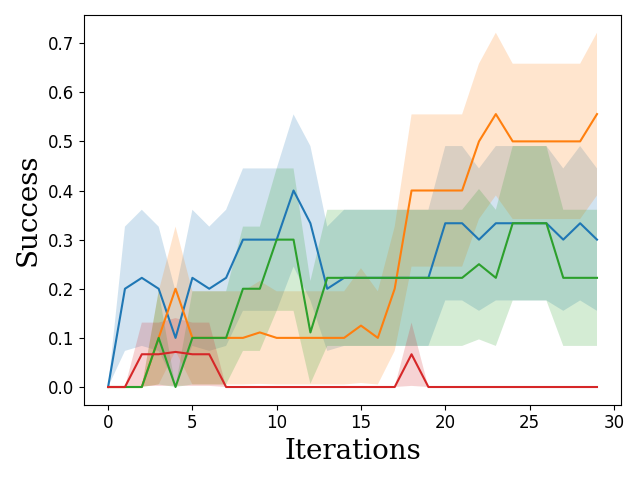}
\caption{Push (Train).}
\end{subfigure}
\\
\begin{subfigure}[t]{0.32\textwidth}
\includegraphics[width=\textwidth]{figures/mwplots/llf-metaworld-reach-v2/Evaluation_success.png}
\caption{Reach (Test)}
\end{subfigure}
\begin{subfigure}[t]{0.32\textwidth}
\includegraphics[width=\textwidth]{figures/mwplots/llf-metaworld-pick-place-v2/Evaluation_success.png}
\caption{Pick-place (Test)}
\end{subfigure}
\begin{subfigure}[t]{0.32\textwidth}
\includegraphics[width=\textwidth]{figures/mwplots/llf-metaworld-push-v2/Evaluation_success.png}
\caption{Push (Test)}
\end{subfigure}
\end{subfigure}

\caption{Learning the feedback control policy (code) for a simulated Sawyer manipulator in LLF-Bench Metawrold. In each iteration (x-axis), one episode of rollout is performed and then the policy is updated. Mean and standard error of success rate over 10 seeds are shown.}
\label{fig:mw_exps full}
\end{figure}

We test the ability of \trace to optimize long-horizon workflows with complex dependencies. We experiment with using \trace to train controller (python code) for a simulated Sawyer robot manipulator.
We use the Meta-World environment of LLF-Bench~\citep{cheng2023llf} as the simulator and consider three tasks reach, pick-place and push. LLF-Bench is a simulated benchmark with gym interface for testing an agent's ability to learn from language feedback.
In these LLF-Bench Meta-World tasks, the observation is a dictionary where each field denotes a feature of the state and has a vector value (e.g., the end-effector position, the goal position, the gripper status, etc.). The keys of the observation dictionary can differ for each task. The action space is 4-dimensional, which controls the relative position of the end-effector and the state of the gripper. In each time step, the LLF-Bench Meta-World simulator returns the observation dictionary and natural language feedback to guide the robot (we use the `a` mode of LLF-Bench, with which the language feedback would contain information about the current performance, explanation of past successes and failures, and suggestions for the next step). An episode ends if the robot successfully solve the problem or because of time-out.
For each task, LLF-Bench also provides a task instruction explaining that the task is about controlling a Sawyer robot arm and the meaning of the action space (see \cite{cheng2023llf}). We use that as the context $\omega$ of the \opto problem.
We consider an episodic setting. For each experiment (a random seed), we randomly sample an initial configuration. Then for each iteration of optimization, we reset the simulator to that sampled initial configuration and run the robot policy for 10\footnote{We set the problem horizon to be 10 steps, as we find the expert policies implemented in LLF-Bench can solve these problems within 10 steps.} steps or until the episode termination due to success. We compute the sum of rewards and gives the output feedback $f$ in texts in the format of ``Success: <true/false> Return: <score>''. Note that the initial condition for all iterations within an experiment is the same so that the optimization problem is deterministic. To evaluate the learned policy's performance, for each experiment, we additionally run the learned policy starting from 10 held-out initial conditions, different from the fixed training initial condition. For each training algorithm discussed, we run it with 30 iterations, where each iteration consists of one episode rollout and one update.

To optimize the controller with \trace, we declare the control code as the parameter using the \bundle decorator with \trainable set to True; the initial control code simply outputs a zero vector $[0,0,0,0]$. We decorate also the reset and the step function of the gym environment, so that the entire rollout of an episode can be traced end-to-end. In our implementation, a prototypical rollout would create a graph with around 30 operations where the controller code parameter is used multiple times. This graph structure is similar to that of running a recurrent neural network.
For \trace, we experiment with \algo with and without a memory of size 10.
In addition to \trace, we implement the state-of-the-art LLM optimizer OPRO~\citep{yang2024large} as a baseline. Compared with \trace, OPRO does not use the execution trace information but rely on just memory of parameter and feedback pairs\footnote{The original version of OPRO uses parameter-score pairs. Since we're interested in the more general setup of \opto, we extend it to use parameter-feedback pairs.
} To make the comparison with \trace fair, we append the feedback observation from LLF-Bench given at each time step to the final feedback received by OPRO; on the other hand, \trace uses the simple final feedback of success and return and has to read the per-step feedback from the execution trace.
To run OPRO in the \opto setting, our implementation only proposes a single candidate in each iteration, which is then evaluated and provided with the output feedback. Since in \citep{yang2024large} OPRO generates about 10 samples per iteration, so one iteration in \citep{yang2024large} is roughly equivalent to 10 iterations here.

The experimental results are summarized in \cref{fig:mw_exps full}, where we show the success rates at both the training initial condition as well as the held-out testing initial conditions over 10 seeds.
\algo is clearly the top-performing optimizer, especially the version with memory. OPRO is able to solve Reach at the start but its performance degraded over iterations (this instability was observed in \citep{yang2024large}) and gets similar performance as \algo (without memory) in Push. To validate that the performance of \algo is indeed due to using the execution trace, we include an ablation where we mask out information in \#Inputs, \#Others, \#Code, \#Definition in the LLM context (see \cref{fig:opt-summary appendix}), which lead to significant degrade in performance and stability. This ablation shows that additionally using the execution trace provides more informed search direction compared with just using just the output feedback, which agrees with our hypothesis.

This experiment features the most complex graph structures, and using \trace for optimization here is similar to back-propagation over time. The experimental results here are quite impressive, showing that \trace is able to learn a complex control logic in a dozens of interactions, not only working on the training initial conditions but also on the held-out testing ones too.
Nonetheless, we want to point out some limitations in the current experimental results. We find that the success rate of the learned policy varies largely across random seeds. Except for Reach (the simplest task), in a seed, often either it finds a policy close to 1.0 success rate or 0.0 success rate. Therefore, the plots can roughly be interpreted as how long it takes to find a working policy. In addition, in these experiments, we find that providing task-related context is necessary. We find the context needs to be informative enough for humans to understand the problem\footnote{The original instructions in the v2 environments of LLF-Bench does not contain task specific background, but only the task name. We find this task name alone, except for Reach, is too vague even for humans to understand how to read the observation dictionary is related to the problem, so we added additionally one line to explain the task. Pick-place: ``The goal of the task is to pick up a puck and put it to a goal position.''. Push: ``The goal of the task is to push a puck to a goal position.''. Without this extra information, none of the LLM-based optimizers works in the experiments.}; otherwise, the optimization can be solved efficiently with the time scale considered here. Nonetheless, this requirement is reasonable, as there is no free lunch.

\clearpage \section{Examples of the Optimized Parameters in the Experiments} \label{sec:optimized params examples}

\begin{figure}[ht]
    \centering
    \begin{lstlisting}[language=Python]
class Policy(trace.Module):
    def init(self, width, height):
        pass

    def __call__(self, map):
        return self.select_coordinate(map).data

    def select_coordinate(self, map):
        plan = self.reason(map)

        act = self.act
        output = act(map, plan)
        return output

    @trace.bundle(trainable=True)
    def act(self, map, plan):
        """
        Given a map, select a target coordinate in a Battleship game. In map, O denotes misses, X denotes successes, and . denotes unknown positions.
        """
        return

    @trace.bundle(trainable=True)
    def reason(self, map) -> str:
        """
        Given a map, analyze the board in a Battleship game. In map, O denotes misses, X denotes successes, and . denotes unknown positions.
        """
        return [0, 0]

def train(epochs=10):
    policy = Policy()
    optimizer = FunctionOptimizerV2Memory(policy.parameters())
    board = BattleshipBoard()

    feedback = ""
    rewards = []
    board = board.reset()
    obs = node(board.get_shots())  # init observation
    i = 0
    while i < epochs:
        try:
            output = policy.select_coordinate(obs)
            # not traced
            obs, reward, terminal, feedback = user_fb_for_placing_shot(board, output.data)  
        except TraceExecutionError as e:  # this is a retry
            output = e.exception_node
            feedback = output.data
            reward, terminal = 0, False

        if terminal:
            board = reset_board()
            obs = node(board.get_shots())  # init observation

        # Update
        optimizer.zero_feedback()
        optimizer.backward(output, feedback)
        optimizer.step(verbose=True)
    
    \end{lstlisting}
    \caption{Battleship Agent and optimization code.}
    \label{fig:battleship-agent}
\end{figure}

\begin{figure}[ht]
    \centering
    \begin{lstlisting}[language=Python]
## Iteration 1 ; Success : False
def reason(self, map) -> str:
    """
    Given a map, analyze the board in a game. On map, O denotes misses, X denotes successes, and . denotes unknown positions.
    """
    # Check for first unknown position to try as a potential move
    for row_index, row in enumerate(map):
        for col_index, val in enumerate(row):
            if val == '.':
                return [row_index, col_index]
    # Default to [0, 0] if no unknown positions found
    return [0, 0]

## Iteration 2 ; Success : True
def reason(self, map) -> str:
    """
    Given a map, analyze the board in a game. On map, O denotes misses, X denotes successes, and . denotes unknown positions.
    """
    # Improved strategic approach by checking . surrounded by more . which might indicate undetected area
    max_surrounding = -1
    best_spot = [0, 0]
    for row_index, row in enumerate(map):
        for col_index, val in enumerate(row):
            if val == '.':
                surrounding = 0
                for di in [-1, 0, 1]:
                    for dj in [-1, 0, 1]:
                        ni, nj = row_index + di, col_index + dj
                        if 0 <= ni < len(map) and 0 <= nj < len(map[0]) and map[ni][nj] == '.':
                            surrounding += 1
                if surrounding > max_surrounding:
                    max_surrounding = surrounding
                    best_spot = [row_index, col_index]
    return best_spot

## Iteration 8 ; Success : True
def reason(self, map) -> str:
    """
    Given a map, analyze the board in a game. On map, O denotes misses, X denotes successes, and . denotes unknown positions.
    """
    max_surrounding = -1
    best_spot = [0, 0]
    for row_index, row in enumerate(map):
        for col_index, val in enumerate(row):
            if val == '.':
                surrounding = 0
                for di in [-1, 0, 1]:
                    for dj in [-1, 0, 1]:
                        ni, nj = row_index + di, col_index + dj
                        if 0 <= ni < len(map) and 0 <= nj < len(map[0]):
                            if map[ni][nj] == '.':
                                surrounding += 1
                            elif map[ni][nj] == 'X':
                                surrounding += 1.5  # increasing emphasis on proximity to successful hits
                if surrounding > max_surrounding:
                    max_surrounding = surrounding
                    best_spot = [row_index, col_index]
    return best_spot
    \end{lstlisting}
    \caption{Learned Battleship Code (reason). Functions with the same name are learned during different iterations or trials.}
    \label{fig:battleship-functions}
\end{figure}

\begin{figure}[ht]
    \centering
    \begin{lstlisting}[language=Python]
## Iteration 0 (Initialization); Success : False


## Iteration 1 ; Success : False
def act(self, map, plan):
    """
    Given a map, select a target coordinate in a game. On map, O denotes misses, X denotes successes, and . denotes unknown positions.
    """
    # Example implementation, more sophisticated logic may be required based on intended gameplay mechanics
    return [0, 1]

## Iteration 2 ; Success : True
def act(self, map, plan):
    """
    Given a map, select a target coordinate in a game. On map, O denotes misses, X denotes successes, and . denotes unknown positions.
    """
    # Use information from plan, which is output from __code2, checking strategic position, not just first available
    best_score = -1
    best_move = [0, 0, 0]  # initially no move is considered
    for row_index, row in enumerate(map):
        for col_index, val in enumerate(row):
            if val == '.':  # consider all unknown spots
                hopes_surrounding = 0  # will hold number of '.'s around the spot
                for di in [-1, 0, 1]:
                    for dj in [-1, 0, 1]:
                        ni, nj = row_index + di, col_index + dj
                        if 0 <= ni < len(map) and 0 <= nj < len(map[0]):
                            if map[ni][nj] == '.':
                                hopes_surrounding += 1
                score = hopes_surrounding  # more '.' around, higher the chance of hit
                if score > best_score:
                    best_score = score
                    best_move = [row_index, col_index]
    if map[plan[0]][plan[1]] == '.' and best_score == -1:
        return plan  # if no better move found, and original place is still unknown
    return best_move if best_score > -1 else plan  # return the best move found, or stick with the plan if still valid

## Iteration 8 ; Success : True
def act(self, map, plan):
    """
    Given a map, select a target coordinate in a game. On map, O denotes misses, X denotes successes, and . denotes unknown positions.
    """
    best_score = -1
    best_move = [0, 0]
    for row_index, row in enumerate(map):
        for col_index, val in enumerate(row):
            if val == '.' and [row_index, col_index] != plan:
                hopes_surrounding = 0
                for di in [-1, 0, 1]:
                    for dj in [-1, 0, 1]:
                        ni, nj = row_index + di, col_index + dj
                        if 0 <= ni < len(map) and 0 <= nj < len(map[0]):
                            if map[ni][nj] == '.':
                                hopes_surrounding += 1  # favoring positions with more unknowns surrounding
                            elif map[ni][nj] == 'X':
                                hopes_surrounding += 2  # increased incentive for moves near successful spots
                score = hopes_surrounding
                if score > best_score:
                    best_score = score
                    best_move = [row_index, col_index]
    if best_score > -1:
        return best_move
    return plan
    \end{lstlisting}
    \caption{Learned Battleship Code (act). Functions with the same name are learned during different iterations or trials.}
    \label{fig:battleship-functions-act}
\end{figure}

\begin{figure}[ht]
    \centering
    \begin{lstlisting}[language=Python]
@trace_class
class Predict(LLMCallable):
    def __init__(self):
        super().__init__()

        self.demos = []
        self.prompt_template = dedent("""
        Given the fields `question`, produce the fields `answer`.

        ---

        Follow the following format.

        Question: 
        Answer: 

        ---
        Question: {}
        Answer:
        """)

        self.prompt_template = trace.node(self.prompt_template, trainable=True,
                                             description="[ParameterNode] This is the Prompt Template to the LLM...")

    @trace.bundle(trainable=True)
    def extract_answer(self, prompt_template, question, response):
        """
        Need to read in the response, which can contain additional thought, delibration and an answer.
        Use code to process the response and find where the answer is.
        Can use self.call_llm("Return the answer from this text: " + response) again to refine the answer if necessary.

        Args:
            prompt_template: The prompt that was used to query LLM to get the response
            question: Question has a text describing the question but also "Options"
            response: LLM returned a string response
                      Process it and return the answer in the exact format that the evaluator wants to see.
                      Be mindful of the type of answer you need to produce.
                      It can be (A)/(B), a number like 8, or a string, or Yes/No.
        """
        answer = response.split("Answer:")[1].strip()
        return answer

    @trace.bundle(trainable=True)
    def create_prompt(self, prompt_template, question):
        """
        The function takes in a question and then add to the prompt for LLM to answer.
        Args:
            prompt_template: some guidance/hints/suggestions for LLM
            question: the question for the LLM to answer
        """
        return prompt_template.format(question)

    def forward(self, question):
        """
        question: text

        We read in a question and produces a response
        """
        user_prompt = self.create_prompt(self.prompt_template, question)
        response = self.call_llm(user_prompt)
        answer = self.extract_answer(self.prompt_template, question, response)
        return answer
    \end{lstlisting}
    \caption{Starting Code for BigBench. We write it in a similar style to DSPy's Predict module.}
    \label{fig:bigbench-predict-class}
\end{figure}

\begin{figure}[ht]
    \centering
    \begin{lstlisting}[language=Python]
@trace_class
class PredictCoT(LLMCallable):
    def __init__(self):
        super().__init__()

        self.demos = []
        self.prompt_template = dedent("""
        Given the fields `question`, produce the fields `answer`.

        ---

        Follow the following format.

        Question: question
        Reasoning: Let's think step by step in order to produce the answer. We ...
        Answer: answer

        ---
        Question: {}
        """)

        self.prompt_template = trace.node(self.prompt_template, trainable=True,
                                             description="[ParameterNode] This is the Prompt Template to the LLM...")

    @trace.bundle(trainable=True)
    def extract_answer(self, prompt_template, question, response):
        """
        Need to read in the response, which can contain additional thought, delibration and an answer.
        Use code to process the response and find where the answer is.
        Can use self.call_llm("Return the answer from this text: " + response) again to refine the answer if necessary.

        Args:
            response: LLM returned a string response
                      Process it and return the answer in the exact format that the evaluator wants to see.
                      Be mindful of the type of answer you need to produce.
                      It can be (A)/(B), a number like 8, or a string, or Yes/No.
            question: Question has a text describing the question but also "Options"
        """
        answer = response.split("Answer:")[1].strip()
        return answer

    @trace.bundle(trainable=True)
    def create_prompt(self, prompt_template, question):
        """
        The function takes in a question and then add to the prompt for LLM to answer.
        The prompt should instruct the LLM to reason, think.
        Args:
            prompt_template: some guidance/hints/suggestions for LLM
            question: the question for the LLM to answer
        """
        return prompt_template.format(question)

    def forward(self, question):
        """
        question: text

        We read in a question and produces a resposne
        """
        user_prompt = self.create_prompt(self.prompt_template, question)
        response = self.call_llm(user_prompt)
        answer = self.extract_answer(self.prompt_template, question, response)
        return answer
    \end{lstlisting}
    \caption{Starting Code for BigBench. We write it in a similar style to DSPy's Predict CoT (0-shot Chain-of-Thought) module.}
    \label{fig:bigbench-predict-cot-class}
\end{figure}

\begin{figure}[ht]
    \centering
    \begin{lstlisting}[language=Python]
## Iteration 0 ( initialization )
def create_prompt(self, prompt_template, question):
    """
    The function takes in a question and then add to the prompt for LLM to answer.
    Args:
        prompt_template: some guidance/hints/suggestions for LLM
        question: the question for the LLM to answer
    """
    return prompt_template.format(question)

## Iteration > 0
def create_prompt(self, prompt_template, question):
    """
    The function takes in a question and then add to the prompt for LLM to answer.
    The prompt should now further instruct the LLM to carefully track the ball swaps occurring step-by-step.
    Args:
        prompt_template: some guidance/hints/suggestions for LLM
        question: the question for the LLM to answer
    """
    prompt_template = 'Process this carefully: Step-by-step.' + prompt_template
    return prompt_template.format(question)
    \end{lstlisting}
    \caption{Learned Predict module for BigBench. Functions with the same name are learned during different iterations or trials.}
    \label{fig:bigbench-functions-p1}
\end{figure}

\begin{figure}[ht]
    \centering
    \begin{lstlisting}[language=Python]
## Iteration 0 ( initialization )
def extract_answer(self, prompt_template, question, response):
    """
    Need to read in the response, which can contain additional thought, delibration and an answer.
    Use code to process the response and find where the answer is.
    Can use self.call_llm("Return the answer from this text: " + response) again to refine the answer if necessary.

    Args:
        prompt_template: The prompt that was used to query LLM to get the response
        question: Question has a text describing the question but also "Options"
        response: LLM returned a string response
                  Process it and return the answer in the exact format that the evaluator wants to see.
                  Be mindful of the type of answer you need to produce.
                  It can be (A)/(B), a number like 8, or a string, or Yes/No.
    """
    answer = response.split("Answer:")[1].strip()
    return answer

## Iteration > 0
def extract_answer(self, response):
    """
    Need to read in the response, which can contain additional thought, deliberation and an answer.
    Use code to process the response and find where the answer is.
    Can use self.call_llm("Return the answer from this text: " + response) again to refine the answer if necessary.
    Args:
        response: LLM returned a string response
                  Process it and return the answer in the exact format that the evaluator wants to see.
                  Be mindful of the type of answer you need to produce.
                  It can be (A)/(B), a number like 8, or a string, or Yes/No.
        question: Question has a text describing the question but also "Options"
    """
    answer = ''
    segments = response.split('\n')
    for segment in segments:
        if 'Answer:' in segment:
            answer = segment.split('Answer:')[1].strip()
    refined_answer = self.call_llm('Return the refined answer from this text: ' + answer)
    return refined_answer

def extract_answer(self, prompt_template, question, response):
    """
    Processes the LLM response and extracts the final answer in the required format.
    """
    # Assuming that the relevant part of the response is after 'Answer:' and before any further commentary
    extracted_part = response.split('Answer: ')[1].split(' ')[0].strip()
    # Find the section of the answer and return it directly
    result = re.search('\([A-E]\)', extracted_part)
    if result:
        return result.group()
    else:
        return 'No valid answer found'

def extract_answer(self, prompt_template, question, response):
    """
    Processes the LLM response, extracting and formatting the final answer.
    Uses code to meticulously parse the response to locate the answer section.
    Optionally refines the answer by querying the LLM again if necessary.

    Args:
        response: string from LLM, expected format contains 'Answer:' followed by the answer.
        question: Description of the question being addressed, may include 'Options'
    """
    answer = response.split('Answer:')[1].split('.')[0].strip().lower().split(',')[0].strip()
    return answer)
    \end{lstlisting}
    \caption{Learned Predict module for BigBench. Functions with the same name are learned during different iterations or trials.}
    \label{fig:bigbench-functions-p2}
\end{figure}

\begin{figure}[ht]
    \centering
    \begin{lstlisting}[language=Python]
## Iteration 0 ( initialization )
prompt_template = """
Given the fields `question`, produce the fields `answer`.

---

Follow the following format.

Question: 
Answer: 

---
Question: {}
Answer:
"""

## Iteration > 0
prompt_template = """
Given the fields `question`, produce the fields `answer`.

---

Follow the following format.

Question: 
Answer: Please respond precisely in the format required.

---
Question: {}
"""
    \end{lstlisting}
    \caption{Learned BigBench Prompt Template.}
    \label{fig:bigbench-template}
\end{figure}

\begin{figure}[ht]
    \centering
    \begin{lstlisting}[language=Python]

## Iteration 0 (initialization); Success: False
def controller(obs):
    """
    A feedback controller that computes the action based on the observation.

    Args:
        obs: (dict) The observation from the environment. Each key is a string (indicating a type of observation) and the value is a list of floats.
    Output:
        action: (list or nd.array) A 4-dimensional vector.
    """
    return [0, 0, 0, 0]

## Iteration 1; Success: False
def controller(obs):
    """
    A feedback controller that computes the action based on the observation.

    Args:
        obs: (dict) The observation from the environment. Each key is a string (indicating a type of observation) and the value is a list of floats.
    Output:
        action: (list or nd.array) A 4-dimensional vector.
    """
    hand_pos = obs['hand_pos']
    goal_pos = obs['goal_pos']
    action = [goal_pos[0] - hand_pos[0], goal_pos[1] - hand_pos[1], goal_pos[2] - hand_pos[2], 0]
    # Adjusting the last element of the action based on whether the hand is close enough to trigger grasping
    if abs(goal_pos[0] - hand_pos[0]) < 0.02 and abs(goal_pos[1] - hand_pos[1]) < 0.02 and abs(goal_pos[2] - hand_pos[2]) < 0.02:
        action[3] = 1 # Assuming action[3] = 1 triggers closing the gripper
    return action

## Interation 5; Success: False
def controller(obs):
    """
    A feedback controller that computes the action based on the observation.

    Args:
        obs: (dict) The observation from the environment.
    Output:
        action: (list or nd.array) A 4-dimensional vector.
    """
    hand_pos = obs['hand_pos']
    puck_pos = obs['puck_pos']
    goal_pos = obs['goal_pos']
    gripper_open = 1 if obs['gripper_distance_apart'][0] > 0.5 else 0

    # Adjusting strategy to close the gripper when in proximity of the puck
    if gripper_open and ((abs(hand_pos[0] - puck_pos[0]) < 0.05) and (abs(hand_pos[1] - puck_pos[1]) < 0.05) and (abs(hand_pos[2] - puck_pos[2]) < 0.05)):
        action = [0, 0, 0, 1]  # Close the gripper
    elif not gripper_open:
        direction_to_goal = [0.09 - hand_pos[0], 0.95 - hand_pos[1], 0.12 - hand_pos[2]]
        action = [direction_to_goal[0], direction_to_goal[1], direction_to_goal[2], 0]  # Move towards the suggested pose once puck is grasped
    else:
        # Move towards the puck first if not carrying it
        direction_to_puck = [puck_pos[0] - hand_pos[0], puck_pos[1] - hand_pos[1], puck_pos[2] - hand_pos[2]]
        action = [direction_to_puck[0], direction_to_puck[1], direction_to_puck[2], 0]

    return action

\end{lstlisting}
    \caption{Learned Code for LLFBench Meta-World Pick-Place (Part 1). Functions with the same name are learned during different iterations or trials.}
    \label{fig:meta-world-pick-place-p1}
\end{figure}

\begin{figure}[ht]
    \centering
    \begin{lstlisting}[language=Python]


## Iteration 10; Success: False
def controller(obs):
    """
    A feedback controller that computes the action based on the observation.

    Args:
        obs: (dict) The observation from the environment.
    Output:
        action: (list or nd.array) A 4-dimensional vector.
    """
    hand_pos = obs['hand_pos']
    puck_pos = obs['puck_pos']
    goal_pos = obs['goal_pos']
    gripper_open = 1 if obs['gripper_distance_apart'][0] > 0.5 else 0

    normalize = lambda x: [i / max(abs(max(x, key=abs)), 1) for i in x]

    # Close the gripper when close to the puck and the gripper is open
    if gripper_open and ((abs(hand_pos[0] - puck_pos[0]) < 0.05) and (abs(hand_pos[1] - puck_pos[1]) < 0.05) and (abs(hand_pos[2] - puck_pos[2]) < 0.05)):
        return [0, 0, 0, 1]
    # When the puck is grasped, move towards the goal position with a normalized direction
    elif not gripper_open and ((abs(hand_pos[0] - puck_pos[0]) < 0.1) and (abs(hand_pos[1] - puck_pos[1]) < 0.1) and (abs(hand_pos[2] - puck_pos[2]) < 0.1)):
        direction_to_goal = [goal_pos[0] - hand_pos[0], goal_pos[1] - hand_pos[1], goal_pos[2] - hand_pos[2]]
        return normalize(direction_to_goal) + [0]
    # When the gripper is open and not close enough to the puck, move towards the puck
    else:
        direction_to_puck = [puck_pos[0] - hand_pos[0], puck_pos[1] - hand_pos[1], puck_pos[2] - hand_pos[2]]
        return normalize(direction_to_puck) + [0]

## Iteration 13; Success: True
def controller(obs):
    """
    A feedback controller that computes the action based on the observation.

    Args:
        obs: (dict) The observation from the environment.
    Output:
        action: (list or nd.array) A 4-dimensional vector.
    """
    hand_pos = obs['hand_pos']
    puck_pos = obs['puck_pos']
    goal_pos = obs['goal_pos']
    gripper_open = 1 if obs['gripper_distance_apart'][0] > 0.5 else 0

    normalize = lambda x: [i / max(abs(max(x, key=abs)), 1) for i in x]
    # Close the gripper when close to the puck and the gripper is open
    if gripper_open and ((abs(hand_pos[0] - puck_pos[0]) < 0.05) and (abs(hand_pos[1] - puck_pos[1]) < 0.05) and (abs(hand_pos[2] - puck_pos[2]) < 0.05)):
        action = [0, 0, 0, 1]  # Close the gripper
    elif not gripper_open and ((abs(hand_pos[0] - puck_pos[0]) < 0.1) and (abs(hand_pos[1] - puck_pos[1]) < 0.1) and (abs(hand_pos[2] - puck_pos[2]) < 0.1)):
        # When the puck is grasped, adjust direction towards the goal with improved precision and ensure successful pickup
        direction_to_goal = normalize([goal_pos[0] - hand_pos[0], goal_pos[1] - hand_pos[1], goal_pos[2] - hand_pos[2]])
        action = direction_to_goal + [1]  # Keep the gripper closed
    else:
        # Move towards the puck first if not carrying it
        direction_to_puck = normalize([puck_pos[0] - hand_pos[0], puck_pos[1] - hand_pos[1], puck_pos[2] - hand_pos[2]])
        action = direction_to_puck + [0]  # Open the gripper to prepare for grasping

    return action
\end{lstlisting}
    \caption{Learned Code for LLFBench Meta-World Pick-Place (Part 2). Functions with the same name are learned during different iterations or trials.}
    \label{fig:meta-world-pick-place-p2}
\end{figure}

\begin{figure}[ht]
    \centering
    
\begin{lstlisting}[escapechar=!]
 # Iteration 1
 { 
    "reasoning": "The task involves controlling a robotic arm to perform a pick-place operation by interpreting observations and generating appropriate control actions (4D vectors) to move the arm towards the goal. The objective is to adjust the controller's !code! (__code0) based on !feedback! received at each step of the process. The !feedback! within #Others suggests whether the movements made by the arm are leading it closer to or farther from the goal, along with advice on what coordinates to aim for next (e.g., [-0.07  0.68  0.12  0.  ]). The controller's !code! currently returns a static action [0, 0, 0, 0], which does not adjust based on observations and hence does not efficiently move the robotic arm towards the goal. Given the repetitive !feedback! suggesting to go to [-0.07  0.68  0.12  0.  ], it appears that the controller needs to dynamically compute actions based on the difference between the current positions (hand_pos, puck_pos) and the goal_pos, along with possibly manipulating the gripper state for pick and place operation.",
    "answer": "",
    "suggestion": {...}
}


# Iteration 5 
 {
    "reasoning": "The !instruction! involves creating a controller for a Sawyer robot arm to solve a pick-and-place task by providing sequential control actions based on observations of the environment and state of the robot. The !feedback! indicates that despite the actions taken, the success criteria have not been met, with the latest attempt achieving a return of 0.9574427394250022. The !code! provided for __code0 defines a controller function using the observed state to compute control actions aimed at grasping a puck and moving it to a goal position while adjusting the gripper based on its distance from the puck. Feedback throughout the execution indicates that the controller adapts its strategy based on proximity to the puck and whether it has been picked up. Feedback also suggests that the gripper's opening and closing actions are central to the task's success, with corrections being made to close the gripper when necessary. The controller's logic for determining movements toward the puck or goal and adjusting the gripper state appears reasonable, but the !feedback! implies that adjustments might be required to ensure task completion. This could include fine-tuning the conditions under which the gripper is closed or adjusting the thresholds for puck pickup and delivery to the goal.",
    "answer": "",
    "suggestion": {}

# Iteration 10 
 {
    "reasoning": "The task involves programming a robot arm to pick up a puck and place it at a goal position by providing sequential 4-dimensional action vectors. The action vectors represent target positions and gripper states (open or close) based on the robot's and puck's current positions. The !feedback! indicates that despite optimizing the controller logic to close the gripper when near the puck, move towards the puck when it's not grasped, and head towards the goal once the puck is grasped, the actions have not resulted in task completion. Specifically, repeated suggestions to close the gripper (indicating it remained open when it should have been closed) and guidance on where to move next suggest that the action selection logic might not be effectively responding to the environment's state or might not be properly engaging the gripper at the right times. The latest !code! includes a condition to normalize the direction of movement and checks for gripper position, but there seems to be a misunderstanding or oversight in adjusting the robot's behavior based on the scenario to successfully complete the pick-place task.",
    "answer": "",
    "suggestion": {...}
}
}



\end{lstlisting}
\caption{Example Responses of LLM for LLFBench Meta-World Pick-Place (Part 1); Content in suggestion is omitted for space (represented as ...); please see the code examples.}
\end{figure}

\begin{figure}[ht]
    \centering
    
\begin{lstlisting}[escapechar=!]
# Iteration 12 
 {
    "reasoning": "In this task, the controller function for the Sawyer robot arm manipulates the robot's actions based on the observed environment to achieve a pick-and-place task. The function correctly defines actions based on the robot's hand position, the puck's position, the goal position, and the gripper's state. However, the !feedback! after each step in the !documentation! indicates that although the robot is making progress towards the goal, there is an issue with the gripper's action state not being switched to 'close' (1) at the crucial moment when the robot is close enough to the puck. Most !feedback! comments suggest that the robot is moving in the right direction but fails to grasp the puck due to the gripper not closing. This inefficiency appears to arise from the conditions set for closing the gripper and possibly the normalized direction towards the goal after the puck has been grasped. The condition for closing the gripper (when close enough to the puck) seems to be met, but the action selected afterwards does not effectively lead to puck manipulation or goal achievement in the most optimized manner, as indicated by the 'Feedback' messages suggesting movements that imply needing adjustments in action selection.",
    "answer": "",
    "suggestion": {...}
}

# Iteration 13 (Success)
LLM response:
 {
    "reasoning": "The !instruction! requests improvements based on !feedback! for a Sawyer robot arm's controller function, aiming to accomplish a pick-and-place task. The controller function accepts observations about the robot and the environment and computes an action to move the robot accordingly. \n\nThe !feedback! provided indicates the task was successfully completed with a suggestion that the final command should have been `[0.04, 0.88, 0.24, 1]`, which probably indicates the precise position and gripper state (closed) needed to achieve the task goal. The !feedback! suggests the need for fine-tuning actions especially towards the end of the task to ensure the robot's gripper is appropriately positioned and the gripper state is correct for completing the task. Since the !feedback! for the current !code! positively notes 'Success: True' with a high return value, no modifications are recommended for `__code0`. \n\nPrevious !feedback! instances with lower success metrics and recommendations for change highlight the iterative improvements to the controller logic, specifically related to grip control and movement precision. The aim was to adjust the controller's response to observation !inputs! to accurately position the gripper and handle the puck. The current !code! evidently manages these aspects well by normalizing directions and managing the gripper's open-close state effectively under varying circumstances (proximity to puck or goal and whether the puck is already grasped). These adjustments have likely led to the successful output in the described problem.",
    "answer": "No changes are recommended as the !feedback! indicates successful task completion and suggests the controller's current logic adequately addresses the task's requirements.",
    "suggestion": {}
}

\end{lstlisting}
\caption{Example Responses of LLMs for LLFBench Meta-World Pick-Place (Part 2); Content in suggestion is omitted for space (represented as ...); please see the code examples.}
\end{figure}

\clearpage
\section*{NeurIPS Paper Checklist}

\begin{enumerate}

\item {\bf Claims}
    \item[] Question: Do the main claims made in the abstract and introduction accurately reflect the paper's contributions and scope?
    \item[] Answer: \answerYes{} 
    \item[] Justification: The contributions (\trace, \opto, \algo) are clearly stated in the introduction and detailed in Sections~\ref{sec:opto},\ref{sec:design},\ref{sec:optimizer}, and claimed improvements are validated in Section~\ref{sec:expts}. Meanwhile the aspirational goal of self-improving interactive agents is elaborated separately in the concluding Section~\ref{sec:conclusion}.

\item {\bf Limitations}
    \item[] Question: Does the paper discuss the limitations of the work performed by the authors?
    \item[] Answer: \answerYes{}
    \item[] Justification: The limitations are discussed in Section~\ref{sec:limitations}.

\item {\bf Theory Assumptions and Proofs}
    \item[] Question: For each theoretical result, does the paper provide the full set of assumptions and a complete (and correct) proof?
    \item[] Answer: \answerYes{} 
    \item[] Justification: There are theorems about the computational complexity of \trace and an information theoretic lower bound on propagated feedback. Both theorems are proved in~\cref{sec:trace analysis}.
    
\item {\bf Experimental Result Reproducibility}
    \item[] Question: Does the paper fully disclose all the information needed to reproduce the main experimental results of the paper to the extent that it affects the main claims and/or conclusions of the paper (regardless of whether the code and data are provided or not)?
    \item[] Answer: \answerYes{} 
    \item[] Justification: Experimental setups are outlined in Section~\ref{sec:expts} and detailed in~\cref{sec:expt_details}. Supplementary material additionally provides the exact code that was run to produce all results, and only requires user to supply an OpenAI API key. %

\item {\bf Open access to data and code}
    \item[] Question: Does the paper provide open access to the data and code, with sufficient instructions to faithfully reproduce the main experimental results, as described in supplemental material?
    \item[] Answer: \answerYes{} 
    \item[] Justification: All of the code to implement \trace and \algo are provided in the supplementary material. All of the experiments described in the paper are reproducible with the supplied code.

\item {\bf Experimental Setting/Details}
    \item[] Question: Does the paper specify all the training and test details (e.g., data splits, hyperparameters, how they were chosen, type of optimizer, etc.) necessary to understand the results?
    \item[] Answer: \answerYes{}
    \item[] Justification: Experimental setups are outlined in Section~\ref{sec:expts} and detailed in~\cref{sec:expt_details}. 

\item {\bf Experiment Statistical Significance}
    \item[] Question: Does the paper report error bars suitably and correctly defined or other appropriate information about the statistical significance of the experiments?
    \item[] Answer: \answerYes{} 
    \item[] Justification: All experiments include standard error bars from $\ge 10$ replications.
    
\item {\bf Experiments Compute Resources}
    \item[] Question: For each experiment, does the paper provide sufficient information on the computer resources (type of compute workers, memory, time of execution) needed to reproduce the experiments?
    \item[] Answer: \answerYes{} %
    \item[] Justification: All the experiments use the same compute resources (LLM API for \algo and machine to run \trace), which are listed at the start of Section~\ref{sec:expts}.
            
\item {\bf Code Of Ethics}
    \item[] Question: Does the research conducted in the paper conform, in every respect, with the NeurIPS Code of Ethics \url{https://neurips.cc/public/EthicsGuidelines}?
    \item[] Answer: \answerYes
    \item[] Justification: The contributions of this paper (a framework analogous to PyTorch to generalize the backpropagation algorithm) do not have direct safety or security implications. 

\item {\bf Broader Impacts}
    \item[] Question: Does the paper discuss both potential positive societal impacts and negative societal impacts of the work performed?
    \item[] Answer: \answerYes{} %
    \item[] Justification: In~\cref{sec:limitations,sec:conclusion} we discuss both the aspirational goals and their broader impacts. 

\item {\bf Safeguards}
    \item[] Question: Does the paper describe safeguards that have been put in place for responsible release of data or models that have a high risk for misuse (e.g., pretrained language models, image generators, or scraped datasets)?
    \item[] Answer: \answerNA{}
    \item[] Justification: There are no models, data or APIs released along with the paper.

\item {\bf Licenses for existing assets}
    \item[] Question: Are the creators or original owners of assets (e.g., code, data, models), used in the paper, properly credited and are the license and terms of use explicitly mentioned and properly respected?
    \item[] Answer: \answerYes{}
    \item[] Justification: Datasets used in the experiments (e.g. BigBench-Hard) are properly attributed.
    
\item {\bf New Assets}
    \item[] Question: Are new assets introduced in the paper well documented and is the documentation provided alongside the assets?
    \item[] Answer: \answerYes{}
    \item[] Justification: There is code implementing \trace and \algo that are released along with the paper. The code repository contains licenses and several tutorial notebooks documenting each functionality. There are no datasets or models in the release, hence datasheets and modelsheets are not applicable for this release.
    
\item {\bf Crowdsourcing and Research with Human Subjects}
    \item[] Question: For crowdsourcing experiments and research with human subjects, does the paper include the full text of instructions given to participants and screenshots, if applicable, as well as details about compensation (if any)? 
    \item[] Answer: \answerNA{}
    \item[] Justification: There are no crowdsourcing or human subject studies conducted in this paper.

\item {\bf Institutional Review Board (IRB) Approvals or Equivalent for Research with Human Subjects}
    \item[] Question: Does the paper describe potential risks incurred by study participants, whether such risks were disclosed to the subjects, and whether Institutional Review Board (IRB) approvals (or an equivalent approval/review based on the requirements of your country or institution) were obtained?
    \item[] Answer: \answerNA{}
    \item[] Justification: The paper does not involve crowdsourcing nor research with human subjects.

\end{enumerate}

\end{document}